\documentclass[a4paper,fleqn]{cas-sc}
\usepackage{makecell}
\usepackage{amsmath}
\usepackage{bm}
\usepackage{pifont}
\usepackage{algorithm}  
\usepackage{algorithmic}
\usepackage{threeparttable}

\usepackage[numbers, compress]{natbib}

\def\tsc#1{\csdef{#1}{\textsc{\lowercase{#1}}\xspace}}
\tsc{WGM}
\tsc{QE}

\begin{document}
\let\WriteBookmarks\relax
\def\floatpagepagefraction{1}
\def\textpagefraction{.001}

\shorttitle{Invertible Koopman neural operator for data-driven modeling of partial differential equations}

\shortauthors{Yuhong Jin et~al.}

\title [mode = title]{Invertible Koopman neural operator for data-driven modeling of partial differential equations}

\tnotemark[1]

\tnotetext[1]{Supported by the National Key R \& D Program of China (Grant No. 2023YFE0125900), National Natural Science Foundation of China (Grant Nos. 12422213, U244120491, and 12372008), the key research and development project of Heilongjiang Province (Grant No. 2023ZX01A03), and the Stabilization Support Project for Basic Military Research Institutes (Grant No. 03020051).}

%

\author[1]{Yuhong Jin}[type=editor,
      style=chinese
]

\author[1]{Andong Cong}[type=editor,
      style=chinese
]







\author[1]{Lei Hou}[type=editor,
      style=chinese,
      orcid=0000-0003-0271-7323
]

\cormark[1]


\ead{houlei@hit.edu.cn}

\ead[url]{http://homepage.hit.edu.cn/houlei}

\author[2]{Qiang Gao}[type=editor,
      style=chinese
]

\author[2]{Xiangdong Ge}[type=editor,
      style=chinese
]

\author[1]{Chonglong Zhu}[type=editor,
      style=chinese
]

\author[3]{Yongzhi Feng}[type=editor,
      style=chinese
]

\author[4]{Jun Li}[type=editor,
      style=chinese
]
\cormark[1]
\ead{y8a82000@163.com}





\cortext[1]{Corresponding author}


\affiliation[1]{organization={School of Astronautics},
      addressline={Harbin Institute of Technology},
      city={Harbin},
      postcode={150001},
      country={P. R. China}}
\affiliation[2]{organization={AECC Shengyang Engine Research Institute},
      city={Shenyang},
      postcode={110000},
      country={P. R. China}}
\affiliation[3]{organization={Harbin Electric Company Limited},
      city={Harbin},
      postcode={150001},
      country={P. R. China}}
\affiliation[4]{addressline={Harbin FPR Institute Co., Ltd.},
      city={Harbin},
      postcode={150001},
      country={P. R. China}}

\begin{abstract}
      Koopman operator theory is a popular candidate for data-driven modeling because it provides a global linearization representation for nonlinear dynamical systems. However, existing Koopman operator-based methods suffer from shortcomings in constructing the well-behaved observable function and its inverse and are inefficient enough when dealing with partial differential equations (PDEs). To address these issues, this paper proposes the Invertible Koopman Neural Operator (IKNO), a novel data-driven modeling approach inspired by the Koopman operator theory and neural operator. IKNO leverages an Invertible Neural Network to parameterize observable function and its inverse simultaneously under the same learnable parameters, explicitly guaranteeing the reconstruction relation, thus eliminating the dependency on the reconstruction loss, which is an essential improvement over the original Koopman Neural Operator (KNO). The structured linear matrix inspired by the Koopman operator theory is parameterized to learn the evolution of observables' low-frequency modes in the frequency space rather than directly in the observable space, sustaining IKNO is resolution-invariant like other neural operators. Moreover, with preprocessing such as interpolation and dimension expansion, IKNO can be extended to operator learning tasks defined on non-Cartesian domains. We fully support the above claims based on rich numerical and real-world examples and demonstrate the effectiveness of IKNO and superiority over other neural operators.
\end{abstract}


\begin{highlights}
      \item A novel data-driven modeling method for PDEs, IKNO, is developed.
      \item INN is introduced to eliminate dependency on reconstruction loss.
      \item Koopman operator is parameterized in frequency space to ensure resolution-invariance.
      \item By preprocessing, such as interpolation, IKNO is available for non-Cartesian domains.
      \item In various numerical and real-world examples, IKNO performs over FNO and KNO.
\end{highlights}

\begin{keywords}
      \sep Deep learning \sep Invertible neural network \sep Koopman operator \sep Data-driven modeling \sep Neural operator \sep Partial differential equations
\end{keywords}

\maketitle

\section{Introduction}

Complex nonlinear dynamical systems are ubiquitous in many engineering fields, such as aerospace and vibration control, and modeling these systems is an important research topic \cite{REN2025119023, XU2025109961, MENG2025118674}. Traditional knowledge-driven modeling approaches usually use a priori expertise to build a set of differential or algebraic equations to describe or explain phenomena of interest, having achieved relative maturity. However, in many scenarios, some key parameters, even expressions of systems of concern, may be difficult to measure or give accurately, making establishing a physical model that can accurately characterize systems' evolution challenging. In recent years, as big data technology and computer performance have improved, data-driven modeling approaches have gained extensive attention from researchers, providing a feasible route to solve the aforementioned problems \cite{WANG2023109242, northReviewDataDrivenDiscovery2023, yeOverviewDataDrivenModels2025}.

Data-driven modeling aims to build surrogate models that capture a system's intrinsic physical pattern based on historical data obtained through measurement or simulation without relying on prior knowledge. Starting from this perspective, researchers have developed a series of data-driven modeling approaches, and representative works include neural networks-based methods \cite{Yang_2023, 9741840, HU2021729}, Sparse Identification of Nonlinear Dynamics (SINDy) \cite{SINDy, SINDyc, naozukaSINDySAFrameworkEnhancing2022, doi:10.1073/pnas.1906995116}, spectral submanifold-based method \cite{cenedeseDatadrivenModelingPrediction2022, 10160418, axasFastDatadrivenModel2023}, and Koopman theory-based methods \cite{9037083, doi:10.1137/18M1177846, DeepKoopman_Lusch, gao2025sparseidentificationnonlineardynamics, wilsonKoopmanOperatorInspired2023}. Among them, the Koopman operator theory-based methods have attracted increasing attention for providing a global linearization representation of nonlinear dynamical systems. Specifically, the Koopman operator defines an infinite-dimensional linear operator acting on a Hilbert space spanned by the observable functions, even if the original system is essentially nonlinear \cite{koopmanHamiltonianSystemsTransformation1931}. Moreover, its finite-dimensional approximation yields a linear matrix, and the Dynamic Mode Decomposition (DMD) is a popular method to obtain it \cite{H_Tu_2014, doi.org/10.1017/S0022112009992059}. The classical DMD method has also been further improved by researchers, giving rise to a series of variants, including Extended DMD (EDMD) \cite{EDMD}, which introduces a nonlinear basis function as a dictionary to solve the finite-dimensional approximation of the Koopman operator in a nonlinear subspace of the observables; Kernel DMD (KDMD) \cite{o.williamsKernelbasedMethodDatadriven2015}, obtained by employing kernel tricks to formalize the EDMD further; Hankel DMD \cite{Arbabi_2017}, which introduces time-delay coordinates in state snapshot pairs; Higher-Order DMD (HODMD) \cite{leclaincheHigherOrderDynamic2017}, established by introducing a sliding window in the time-delay form of state snapshots and sequentially employing classical DMD in the window; multiresolution DMD (mrDMD) \cite{doi:10.1137/15M1023543}, that integrates multiresolution analysis and DMD to enable better handling of complex dynamical systems at multiple scales; sparsity-promoting DMD (spDMD) \cite{jovanovicSparsitypromotingDynamicMode2014} with additional consideration of regularization concerning the model amplitudes, which ensures that the model amplitudes have a sparse structure, and thus achieves a more desirable trade-off between the quality of the reconstruction and the number of modes; de-biased DMD \cite{hematiDebiasingDynamicMode2017}, which eliminates bias error through an augmented state snapshot matrix in the subspace projection, thus enhancing robustness; forward-backward DMD (fbDMD) \cite{dawsonCharacterizingCorrectingEffect2016}, which additionally takes into account the system's inverse dynamics and corrects for the bias caused by noises; and physics-informed DMD \cite{baddooPhysicsinformedDynamicMode2023a} with integrating the system's physical information on top of DMD, such as conservation, self-adjointness, and shift-equivariance, to further constrain the form of the linear matrix adopted to approximate the Koopman operator. Furthermore, Haller and Kasz{\'a} \cite{hallerDatadrivenLinearizationDynamical2024} mathematically rigorously prove and clarify some of the limitations and underlying assumptions of DMD. Based on this, a novel advanced strategy for DMD is developed by incorporating the spectral submanifold. Haseli and Cort{\'e}s et al. \cite{haseliLearningKoopmanEigenfunctions2022} point out that information will be lost when the dictionary adopted by the EDMD method does not span the Koopman operator's invariant subspace, leading to inapplicability to the long-term prediction. Subsequently, they derive the necessary and sufficient  conditions for identifying eigenfunctions based on the dictionary functions and propose the Symmetric Subspace Decomposition (SSD) algorithm, which can solve all the Koopman eigenfunctions and span them into the Koopman operator's maximal invariant subspace from any dictionary functions, significantly improving EDMD's performance. The above methods have also achieved outstanding results in various engineering applications \cite{liPredictionWindEnergy2024, endoManifoldAlterationMajor2024, lortieAsymptoticallyStableDataDriven2024, leventidesExtendedDynamicMode2022}. Mendez et al. \cite{mendezNewAutomaticVery2021} establish an efficient approach for analyzing aircraft flutter test data based on HODMD and utilize it to analyze flight test data provided by Gulfstream. The results show that the proposed approach can reconstruct the spatial distribution of aircraft surface modes and provide helpful information for flutter prediction. Ma et al. \cite{maAdaptiveDynamicMode2022} use DMD to decompose the vibration signals of rolling bearings, and the related experimental results show that compared with Variational Mode Decomposition (VMD) and other techniques, the envelope spectra of the model components obtained by DMD can better reflect the fault characteristics.

Giving well-behaved observable functions is important for obtaining the finite-dimensional approximation of the Koopman operator \cite{iacobKoopmanFormNonlinear2024, renPredictionSpatiotemporalDynamic2024}. Moreover, to predict the original system's state, constructing the inverse corresponding to the observable functions is often necessary \cite{doi:10.1137/18M1177846, 9197510}. In traditional approaches, the observation functions are generated based on the given basis functions, such as monomials, Hermite polynomials, Fourier basis functions, etc. \cite{10091950, 9442852, klusDatadrivenApproximationKoopman2020, hanDatadrivenKoopmanModeling2025, 9277915}, whose representation capability is limited. Hence, it has become a trend to integrate the Koopman operator theory with some other advanced strategies, such as SINDy \cite{wangImprovedKoopmanMPCFramework2023, luDeepEmbeddingKoopman2024, luFlightKoopmanDeepKoopman2025} and neural networks \cite{liExtendedDynamicMode2017b, MAKSAKOV2023110368, leaskModalExtractionSpatiotemporal2021a, alford-lagoDeepLearningEnhanced2022, 9968056}. Lu et al. \cite{luDeepEmbeddingKoopman2024} constructs the observation functions in time-delay coordinates via SINDy and proposes a novel approach for predicting nonlinear flight training trajectories. Wang et al. \cite{wangPhysicsinformedDeepKoopman2024} adopt the Lagrangian neural network \cite{cranmer2020lagrangianneuralnetworks} to parameterize the observation functions and take energy difference matching into account during the training process, thus introducing physical constraints with explicit meaning, i.e., Lagrangian dynamics. Zhang et al. \cite{zhangLearningHamiltonianNeural2024} develop the Hamiltonian Neural Koopman Operator (HNKO) for the Hamiltonian system, which adopts an auto-encoder structured neural network to learn the observable function and its inverse. Simultaneously, similar to reference \cite{baddooPhysicsinformedDynamicMode2023a}, they further constrain the linear matrix for approximating the Koopman operator is orthogonal to sustain and discover potential conservation laws. Meng et al. \cite{10682797} and Tayal et al. \cite{tayal2023koopman} are motivated by similar reasons to improve the work of Azencot et al. \cite{azencot2020forecasting} based on neural networks. Utilizing an approach similar to fbDMD, they show that the system's inverse dynamics should be additionally considered when solving the finite-dimensional approximation of the Koopman operator. The results show that such a treatment can effectively enhance the consistency of the Koopman operator. Different from above previous reports, our works \cite{JIN2023110604, jinExtendedDynamicMode2024} point out and formally formulate the reversibility problem that exists in the construction of observation function and its inverse based on the auto-encoder structured neural network for the first time and introduce the Invertible Neural Network to address this issue. Other researchers' subsequent works demonstrate similar motivations \cite{houInvertibleNeuralNetwork2024a, meng2023physicsinformed, li2023invka}. 

However, the aforementioned Koopman operator-based methods are usually only applicable to systems described by ordinary differential equations. Although examples of Partial Differential Equations (PDEs) are reported in some works, all of them require rearrangement of the PDE's spatial states divided by meshes into high-dimensional vectors \cite{renPredictionSpatiotemporalDynamic2024, curtisMachineLearningEnhanced2023, KORDA2018149} or supplementing with dimensionality reduction techniques, such as Proper Orthogonal Decomposition (POD) \cite{doi:10.1137/18M1177846, wilsonKoopmanOperatorInspired2023}, leading to objective limitations. In recent years, operator learning and the neural operator have emerged as a powerful methodology for solving data-driven modeling problems for PDEs, resulting in many outstanding works, and a very partial list include Deep Operator Network (DeepONet) \cite{Lu_2021} and its variants \cite{LI2025128675, chenHybridDecoderDeepONetOperator2024, venturiSVDPerspectivesAugmenting2023, GOSWAMI2024116674, doi:10.1137/23M1598751}, multipole neural operator \cite{li2020multipolegraphneuraloperator}, graph neural operator \cite{li2020neuraloperatorgraphkernel}, Laplace neural operator \cite{caoLaplaceNeuralOperator2024}, wavelet-based neural operator \cite{li2024m2nomultiresolutionoperatorlearning, guptaMultiwaveletbasedOperatorLearning2021, TRIPURA2023115783}, convolution neural operator \cite{raonicConvolutionalNeuralOperators2023b}, Transformer-based neural operator \cite{cao2021choosetransformerfouriergalerkin}, and the very popular Fourier Neural Operator (FNO) \cite{li2021fourierneuraloperatorparametric} and its variants \cite{guibasAdaptiveFourierNeural2022, pathakFourCastNetGlobalDatadriven2022, rahmanUshapedNeuralOperators2023, WEN2022104180, youLearningDeepImplicit2022}. These methods have also been applied to various engineering applications, such as airfoil flow \cite{mengFastFlowPrediction2023} and urban microclimate \cite{PENG2024111063}. The neural operator theory lays the foundation for further extending the application boundaries of the Koopman operator theory, especially for PDEs. From this perspective, Xiong et al. \cite{XIONG2024113194, xiongKoopmanLabMachineLearning2023} propose the Koopman Neural Operator (KNO), which integrates the Koopman operator theory and neural operator. Moreover, Meng et al. \cite{mengKoopmanNeuralOperator2024a} and Cao et al. \cite{CAO2024544} apply the KNO to construct surrogate models for predicting the transonic airfoil flow field and stress-released distortion in large blade machining, demonstrate KNO's effectiveness. However, the original KNO still constructs the observable function and inverse via an auto-encoder and relies on additional reconstruction loss for training, which is deficient, as emphasized in our and other researchers' previous works \cite{JIN2023110604, jinExtendedDynamicMode2024, houInvertibleNeuralNetwork2024a, meng2023physicsinformed}.

The motivation of this paper is to develop a novel data-driven modeling method for PDEs based on the Koopman operator theory and neural operator to extend our previous works further \cite{JIN2023110604, jinExtendedDynamicMode2024} and advance the original KNO. An Invertible Neural Network is introduced to simultaneously parameterize the observable function and its inverse with the same trainable parameters, addressing the dependency on reconstruction loss of the auto-encoder structure. The structured linear matrix inspired by the Koopman operator theory is parameterized to learn the evolution of observables' low-frequency modes in the frequency space implemented based on the Fast Fourier Transform and lowpass filtering. This pipeline ensures the proposed method's resolution-invariance like other neural operators. Then, a convolution layer and the activation function are utilized to extract the high-frequency information and mix them with the inverse Fourier transform of the low-frequency modes. Based on this, the prediction results under the original physical space are obtained through the observable function's inverse. Various numerical and real-world examples are discussed, and the corresponding results demonstrate the proposed method's effectiveness and superiority over the FNO and original KNO.

\section{Data-driven modeling of the partial differential equations: problem formulation}

In general, consider a PDE described by
\begin{equation}
      \begin{aligned}
            (\mathcal{L}u)(x, t) + (\mathcal{N}u)(x, t) &= 0, x \in \Omega, t \in [0, T_{max}] \\
            u(x, t) &= u_{bc}(x, t), x \in \partial \Omega, t \in [0, T_{max}] \\
            u(x, 0) &= u_{ic}(x), x \in \Omega
      \end{aligned} 
      \label{eq:PDE}
\end{equation}
where $\mathcal{L}$ is the linear differential operator, $\mathcal{N}$ is the nonlinear differential operator, $\Omega \subset \mathbb{R}^{d_{x}}$ is the spatial domain with boundary $\partial \Omega$. Without losing generality, Eq.\eqref{eq:PDE} is time-dependent, and $T \triangleq [0, T_{max}]$ is known as the time domain. $u_{bc}$ and $u_{ic}$ are the boundary condition and initial condition, respectively. Note that only the Dirichlet boundary condition is shown for simplicity. The Neumann or Robin boundary condition can be similarly considered. Then, consider the PDE given in Eq.\eqref{eq:PDE} is unknown and parameterized by the input function $a \in \mathcal{A} \triangleq \mathcal{B} (\Omega; \mathbb{R}^{d_{a}})$, for example, the varying boundary condition $u_{bc}(x, t): \partial \Omega \times T \mapsto \mathbb{R}$ or the changing initial condition $u_{ic}(x): \Omega \mapsto \mathbb{R}$, where $\mathcal{B}(\cdot; \cdot)$ represents the separable Banach space. Now the corresponding PDE's solution can be denoted as $u(x, t) \in \mathcal{U} \triangleq \mathcal{B} (\Omega \times T; \mathbb{R})$. The mapping between $\mathcal{A}$ and $\mathcal{U}$ is controlled by Eq.\eqref{eq:PDE}. Moreover, define the measurable output function $w \in \mathcal{W} \triangleq \mathcal{B} (\Omega; \mathbb{R}^{d_{w}})$, which can be obtained by the pre-defined output operator $\mathcal{O}: \mathcal{U} \mapsto \mathcal{W}$. The input-output operator can be defined as $\mathcal{J}: \mathcal{A} \mapsto \mathcal{W}$. Based on the above notation, the data-driven modeling problem for the PDE is formulated as: suppose we have the observations of the input-output function pairs $\{a_{i}, w_{i}\}_{i=1}^{N}$, where $a_{i} \in \mathcal{A}$ and $w_{i} \in \mathcal{W}$. The goal is to learn a surrogate operator $\hat{\mathcal{J}}$ parameterized by $\mathbf{\Theta}$ (usually, a neural network) based on the given input-output function pairs, such that $\hat{\mathcal{J}} \approx \mathcal{J}$. Note that $\hat{\mathcal{J}}$ is an operator describing a mapping between two infinite-dimensional function spaces, which differs from the traditional problem.

Specifically, in time-dependent PDEs, we pay additional attention to the problem of prediction on the time scale, i.e., $a(x) = [u(x, t_{0}-(T_{d}-1)\Delta t),\cdots, u(x, t_{0}-\Delta t), u(x, t_{0})]$ and $w(x) = u(x, t_{0}+\Delta t)$, where $\Delta t$ is the time interval, and $T_{d}$ is the window size. This problem can be formulated as $u|_{\Omega \times \{ t_{0}-(T_{d}-1)\Delta t, \cdots, t_{0} \}} \mapsto u|_{\Omega \times \{ t_{0} + \Delta t \}}$. Moreover, iterating the above mapping by autoregression allows for multi-step prediction, denoted as $u|_{\Omega \times \{ t_{0}-(T_{d}-1)\Delta t, \cdots, t_{0} \}} \mapsto u|_{\Omega \times \{ t_{0} + \Delta t, \cdots, t_{0} + T_{p} \Delta t \}}$, where $T_{p}$ is the prediction horizon.

Although $a_{i}$ and $w_{i}$ are continuous functions in the problem formulation, in practice, we operate on them point-wise in the discrete case. Consider $a_{i}$ and $w_{i}$ on the $R = R_{1} \times R_{1} \times \cdots R_{d_{x}}$ grid points, denoted as $\Omega_{0} \triangleq \{x_{j}\}_{j=1}^N \subset \Omega$, where $R_{i}, i = 1, \cdots, d_{x}$ is the number of grid points (known as the resolution) in the $i$-th dimension. In this case the input-output function pairs are finite-dimensional, i.e., $a_{i} |_{\Omega_{0}} \in \mathbb{R}^{R \times d_{a}}$ and $w_{i}|_{\Omega_{0}} \in \mathbb{R}^{R \times d_{w}}$. The surrogate operator $\hat{\mathcal{J}}$ is additionally required to be resolution-invariant or, equivalently, discretization-convergent or mesh-independent. That is, $\hat{\mathcal{J}}$ can output the solution for any $x \in \Omega$, even for $x \notin \Omega_{0}$. For example, $\hat{\mathcal{J}}$ is trained on the resolution $R_{1} \times R_{1} \times \cdots R_{d_{x}}$ and can predict the solution on the resolution $2R_{1} \times 2R_{1} \times \cdots 2R_{d_{x}}$ or higher. This property allows the generalization between different mesh discretizations, which is crucial for the practical application of the data-driven modeling method.

\subsection{Neural operator}

The neural operator provides an effective route to establish $\hat{\mathcal{J}}$. For ease of presentation, we will not introduce additional notation here and will directly denote the neural operator by $\hat{\mathcal{J}}$. Formally, a complete neural operator can be represented as
\begin{equation}
      \hat{\mathcal{J}}: a \mapsto \mathcal{E}(a) = v_{0} \mapsto v_{1} \cdots \mapsto v_{l} \mapsto \mathcal{D}(v_{l}) = \hat{w}
      \label{eq:neural_operator}
\end{equation}
where $\hat{w} \in \mathcal{B}(\Omega; \mathbb{R}^{d_{w}})$ is the predicted output. $\mathcal{E}: a \mapsto \{ v_{0} \in \mathcal{B}(\Omega; \mathbb{R}^{dv}) \}$ represents the encoder, which maps the input function $a$ to the latent space with higher dimension $dv$. In the neural network framework, $\mathcal{E}$ can be instantiated by a shallow multi-layer perceptron (MLP). $v_{0} \mapsto v_{l}$ is accomplished through $l$ numbers of iterations. Let $\mathcal{D}_{i, i-1}: \{ v_{i-1} \in \mathcal{B}(\Omega; \mathbb{R}^{dv}) \} \mapsto \{ v_{i} \in \mathcal{B}(\Omega; \mathbb{R}^{dv}) \}, i = 1, \cdots, l$, represents the $i$-th iteration. Through the kernel integral operator \cite{li2020neuraloperatorgraphkernel}, $\mathcal{D}_{i, i-1}$ has the following form
\begin{equation}
      v_{i} (x) = \mathcal{D}_{i, i-1}(v_{i-1})(x) = \gamma(\int_{\Omega} k(x, y) v_{i-1}(y) dy + \mathcal{H}(v)(x))
      \label{eq:kernel_integral_operator}
\end{equation}
where $k: \mathbb{R}^{2d_{x}} \mapsto \mathbb{R}^{d_{v} \times d_{v}}$ is the kernel function, which can be parameterized by a neural network and learned from the data autonomously. $\mathcal{H}: \mathcal{B}(\Omega; \mathbb{R}^{d_{v}}) \mapsto \mathcal{B}(\Omega; \mathbb{R}^{d_{v}})$ is a linear transformation, which can be achieved by a linear layer or, equivalently, 
a convolutional layer with kernel size $1^{d_{v}}$. $\gamma: \mathbb{R} \mapsto \mathbb{R}$ is the point-wise nonlinear activation function. Specifying the kernel function or $\mathcal{D}_{i, i-1}$ in a different form allows for establishing neural operators with different structures \cite{li2020multipolegraphneuraloperator, li2024m2nomultiresolutionoperatorlearning, WEN2022104180, TRIPURA2023115783, CAO2024544, XIONG2024113194}. Finally, the decoder $\mathcal{D}: \{ v_{l} \in \mathcal{B}(\Omega; \mathbb{R}^{d_{w}}) \} \mapsto \hat{w}$ realized by another shallow MLP maps the latent space back to the output space. 

\subsection{Koopman operator theory}

Consider the following time-discrete, autonomous dynamical system
\begin{equation}
      \mathbf{x}_{n+1} = \mathbf{F}(\mathbf{x}_{n})
      \label{eq:discrete_dynamical_system}
\end{equation}
where $\mathbf{x}_{n} \in \mathbb{R}^{m}$ is the state at time-step $n$, and $\mathbf{F}: \mathbb{R}^{m} \mapsto \mathbb{R}^{m}$ is the nonlinear evolution operator. The Koopman operator $\mathcal{K}$ is defined as an infinite-dimensional linear operator acting on the Hilbert space, and its finite-dimensional approximation yields a linear matrix, given by
\begin{equation}
      \mathbf{g}(\mathbf{x}_{n+1}) = \mathbf{K} \mathbf{g}(\mathbf{x}_{n})
      \label{eq:koopman_operator}
\end{equation}
where $\mathbf{g}: \mathbb{R}^{m} \mapsto \mathbb{C}^{O}$ is the vector-valued observable function, and $\mathbf{K} \in \mathbb{C}^{O \times O}$ is the Koopman matrix. $\mathbf{K}$ and $\mathbf{g}$ jointly inscribe some of the eigenvalues and eigenfunctions of the Koopman operator $\mathcal{K}$ and can be solved in a data-driven manner, such as DMD \cite{H_Tu_2014}, EDMD \cite{williamsDataDrivenApproximation2015}, and currently highly popular neural network-based methods \cite{houInvertibleNeuralNetwork2024a, GARMAEV2024110351, yuDynamicModeDecomposition2025a}. It can be seen from Eq.\eqref{eq:koopman_operator} that the Koopman operator gives a global linearized representation of the nonlinear system. In addition, to predict the original system's evolution, giving the inverse of the $\mathbf{g}$ is usually necessary.

Note that Eq.\eqref{eq:koopman_operator} has the time-delay version, which can be represented as
\begin{equation}
      \mathbf{g}(\mathbf{x}^{aug}_{n+1}) = \mathbf{K}^{aug} \mathbf{g}(\mathbf{x}^{aug}_{n})
      \label{eq:koopman_operator_time_delay}
\end{equation}
where $\mathbf{x}^{aug}_{n} = [\mathbf{x}_{n-s+1}; \cdots; \mathbf{x}_{n-1}; \mathbf{x}_{n}] \in \mathbb{R}^{(n \times s)}$ is the state vector augmented by the time-delay coordinates. Eq.\eqref{eq:koopman_operator_time_delay} is related to the Time-Delay Embedding Theorem \cite{10.1007/BFb0091924} and has also been achieved a wide range of applications \cite{Arbabi_2017, bruntonChaosIntermittentlyForced2017, curtisMachineLearningEnhanced2023}. 

\subsection{Invertible Neural Network}

\begin{figure}[htbp]
      \centering
      \includegraphics[width = 0.99\textwidth]{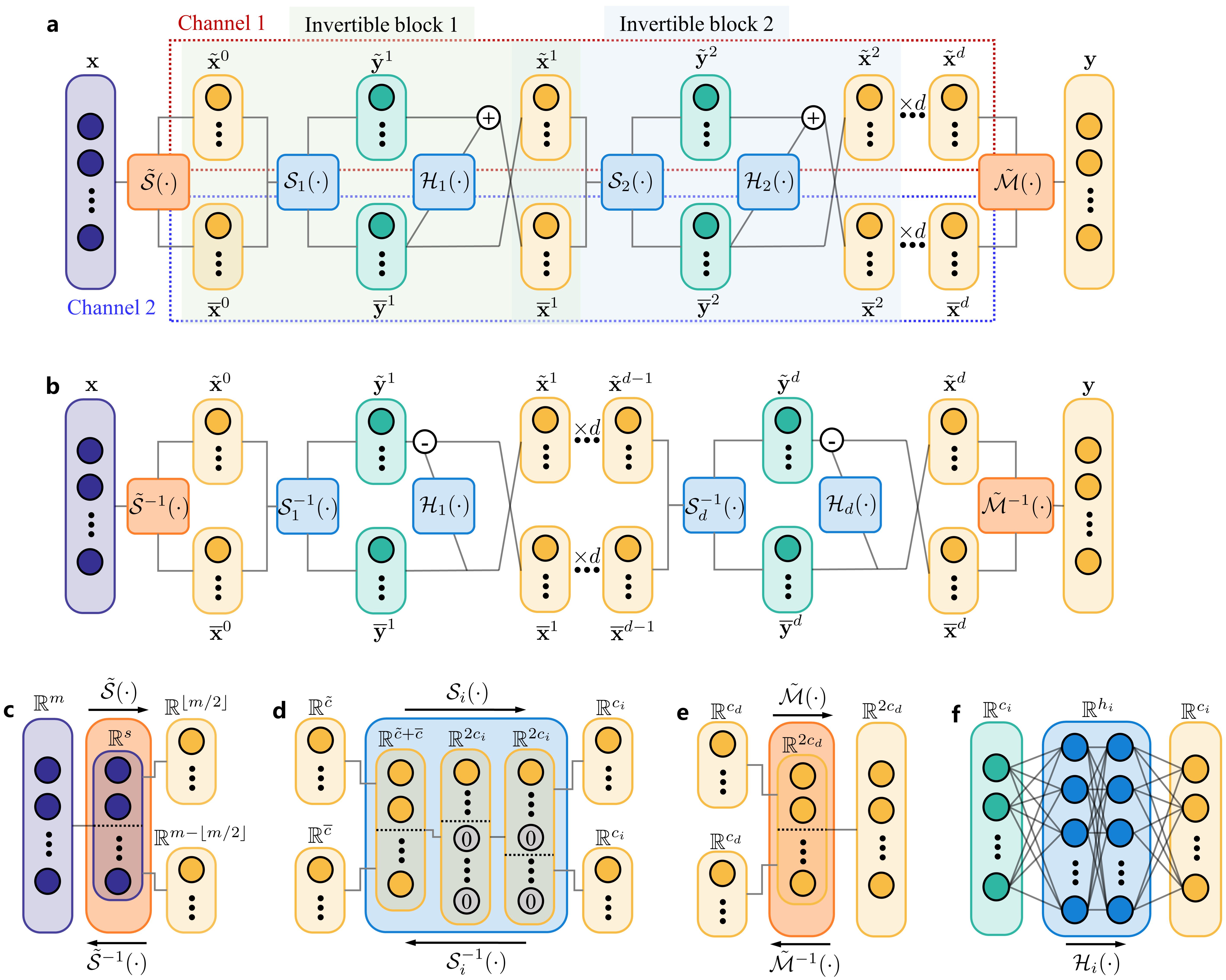}
      \caption{Structure of the INN. \textbf{a} Forward process of the INN. \textbf{b} Inverse process of the INN. \textbf{c} Splitting operation $\tilde{\mathcal{S}}$ and its inverse $\tilde{\mathcal{S}}^{-1}$. \textbf{d} Zeros-concatenating operation $\mathcal{S}_{i}$ and its inverse $\mathcal{S}_{i}^{-1}$. \textbf{e} Merging operation $\tilde{\mathcal{M}}$ and its inverse $\tilde{\mathcal{M}}^{-1}$. (f) Nonlinear mapping operation $\mathcal{H}_{i}$ (irreversible).}
      \label{fig:invertible_neural_network}
\end{figure}

This subsection will briefly introduce a particular type of neural network, the Invertible Neural Network (INN), which is one of the focuses of the subsequent methodology. Unlike general neural networks, the INN has two different computational pipelines under the same set of parameters, i.e., forward and inverse processes. Inspired by the work of Jacobsen et al. \cite{jacobsen2018irevnet}, as shown in Fig.\ref{fig:invertible_neural_network}a, the INN consists of multiple invertible blocks and some invertible dimension operations. Firstly, the input is split into two channels during the forward process, given by
\begin{equation}
      [\tilde{\mathbf{x}}^{0}, \overline{\mathbf{x}}^{0}] = \tilde{\mathcal{S}} (\mathbf{x})
      \label{eq:splitting}
\end{equation}
where $\mathbf{x} \in \mathbb{R}^{m}$, $\tilde{\mathbf{x}}^{0} \in \mathbb{R}^{\lfloor m/2 \rfloor}$, $\overline{\mathbf{x}}^{0} \in \mathbb{R}^{m - \lfloor m/2 \rfloor}$, and $\lfloor \cdot \rfloor$ represents the floor function. In the subsequent representation, superscripts $\tilde{\cdot}$ and $\overline{\cdot}$ are utilized to mark the variables in the two channels, respectively. $\tilde{\mathcal{S}}$ is the splitting operation, as shown in Fig.\ref{fig:invertible_neural_network}c, which can be realized by a simple slicing operation. Obviously, it is invertible and its corresponding inverse is a merging operation, denoted as $\tilde{\mathcal{S}}^{-1}$. Then, there exists a hyperparameter for the $i$th invertible block that becomes the desired dimension $c_{i}$, and satisfying $c_{i} \geqslant c_{i-1} \geqslant \cdots \geqslant c_{d} \geqslant \lceil n_{s}/2 \rceil$, where $\lceil \cdot \rceil$ represents the ceil function. To align dimensions, the zeros-concatenating operation $\mathcal{S}_{i}$, as plotted in Fig.\ref{fig:invertible_neural_network}d, is introduced in each invertible block. Notice that $\mathcal{S}_{i}$ is also invertible as long as the number of padding is recorded, and its inverse is denoted as $\mathcal{S}_{i}^{-1}$. Hence, the computational process of the $i$th invertible block can be described as
\begin{equation}
      \begin{aligned}
            \left[ \tilde{\mathbf{y}}^{i}, \overline{\mathbf{y}}^{i} \right] & = \mathcal{S}_{i}( \left[ \tilde{\mathbf{x}}^{i-1}, \overline{\mathbf{x}}^{i-1} \right] ) \\
            \left[ \tilde{\mathbf{x}}^{i}, \overline{\mathbf{x}}^{i} \right] & = \mathcal{R}_{i} (\left[ \tilde{\mathbf{y}}^{i}, \overline{\mathbf{y}}^{i} \right]) = \left[ \overline{\mathbf{y}}^{i}, \mathcal{H}_{i}(\overline{\mathbf{y}}^{i}) + \tilde{\mathbf{y}}^{i} \right] \\
      \end{aligned}
      \label{eq:invertible_block}
\end{equation}
where $\tilde{\mathbf{y}}^{i}$ and $\overline{\mathbf{y}}^{i}$ are intermediate variables. $\tilde{\mathbf{x}}^{i}$ and $\overline{\mathbf{x}}^{i}$ are the output of the $i$th invertible block. All of them are in the $\mathbb{R}^{c_{i}}$ space. $\mathcal{H}_{i}$ is the nonlinear mapping operation, which can be implemented by a shallow MLP (shown in Fig.\ref{fig:invertible_neural_network}f). Here, we adopt a three-layer MLP with a residual connection to instantiate $\mathcal{H}_{i}$. The hidden dimension is known as $h_{i}$. It is imperative to emphasize that $\mathcal{H}_{i}$ does not need to introduce any additional constraints on the structure other than the requirement that the input and output dimensions are the same, guaranteeing the representation ability of the INN. Subsequently, only half of the features in Eq.\eqref{eq:invertible_block} are nonlinearly transformed. However, this is not a serious problem and can be solved by the flip operation implicit in $\mathcal{R}_{i}$ after stacking two or more layers of invertible blocks. Besides, due to the inclusion of a natural residual structure, Eq.\eqref{eq:invertible_block} is also known as the residual coupling functions \cite{papamakarios2021normalizing, gomez2017reversible}. Finally, the variables in the two channels are combined by a merging operation to get the final output
\begin{equation}
      \mathbf{y} = \tilde{\mathcal{M}}(\tilde{\mathbf{x}}^{d}, \overline{\mathbf{x}}^{d})
      \label{eq:merging}
\end{equation}
where $d$ is the INN's depth. $\tilde{\mathcal{M}}$ is the merging operation, whose inverse is a simple splitting operation, denoted as $\tilde{\mathcal{M}}^{-1}$ (shown in Fig.\ref{fig:invertible_neural_network}e). Based on the above, the INN's forward process can be expressed in a more compact form as
\begin{equation}
      \mathbf{y} = \tilde{\mathcal{M}} \circ \mathcal{R}_{d} \circ \mathcal{S}_{d} \circ \cdots \circ \mathcal{R}_{1} \circ \mathcal{S}_{1} \circ \tilde{\mathcal{S}} (\mathbf{x})
      \label{eq:forward_process}
\end{equation}
where $\circ$ is the composition operation.

Reconsidering Eq.\eqref{eq:invertible_block}, it can be observed that the inverse of $\mathcal{R}_{i}$, $\mathcal{R}_{i}^{-1}$, can be explicitly obtained
\begin{equation}
      \left[ \tilde{\mathbf{y}}^{i}, \overline{\mathbf{y}}^{i} \right] = \mathcal{R}_{i}^{-1} (\left[ \tilde{\mathbf{x}}^{i}, \overline{\mathbf{x}}^{i} \right]) = \left[ \overline{\mathbf{x}}^{i}-\mathcal{H}_{i}(\tilde{\mathbf{x}}^{i}), \tilde{\mathbf{x}}^{i} \right]
      \label{eq:invertible_block_inverse}
\end{equation}
Hence, the INN's inverse process is described as
\begin{equation}
      \mathbf{x} = \tilde{\mathcal{S}}^{-1} \circ \mathcal{S}_{1}^{-1} \circ \mathcal{R}_{1}^{-1} \circ \cdots \circ \mathcal{S}_{d}^{-1} \circ \mathcal{R}_{d}^{-1} \circ \tilde{\mathcal{M}}^{-1} (\mathbf{y})
      \label{eq:inverse_process}
\end{equation}
Remarkably, $\mathbf{x} = \mathcal{G}^{-1}(\mathcal{G}(\mathbf{x}))$ holds strictly for arbitrary identical parameters, which shows that the INN can act as both encoder and decoder without introducing additional parameters and training task. This is a well-behavioral property for constructing the observable function and its corresponding inverse adopted to obtain the Koopman operator's finite-dimensional approximation.

\section{Proposed Invertible Koopman Neural Operator}

This section focuses on providing a detailed explanation for the proposed method, called Invertible Koopman Neural Operator (IKNO), including its components and calculation process. In addition, we also give the mathematical counterparts of the IKNO's each component in the neural operator theory.

\subsection{Components and calculation process}

This subsection details the core components of the proposed IKNO and the change of dimensionality during the operation, using a two-dimensional PDE as an example, which helps intuitivly understand the IKNO's structure. As shown in Fig.\ref{fig:ikno_structure}a, consider learning a neural operator mapping the initial $T_{d}$ snapshots to the snapshots up to same later time, defined by $\hat{\mathcal{J}} \approx \mathcal{J}: u|_{\Omega \times \{ t_{0}-(T_{d}-1)\Delta t, \cdots, t_{0} \}} \mapsto u|_{\Omega \times \{ t_{0} + \Delta t, \cdots, t_{0} + T_{p} \Delta t \}}$. After discrete sampling, the input to the IKNO can be expressed as a fourth-order tensor, $\mathbf{u}^{i}_{0 \thicksim T_{d}-1} \in \mathbb{R}^{B \times R_{1} \times R_{2} \times T_{d}}$. $B$ is the batch size, $R_{1}$ and $R_{2}$ represent the spatial resolution. The core components of the IKNO architecture include the following parts:

\begin{figure}[htbp]
      \centering
      \includegraphics[width = 0.99\textwidth]{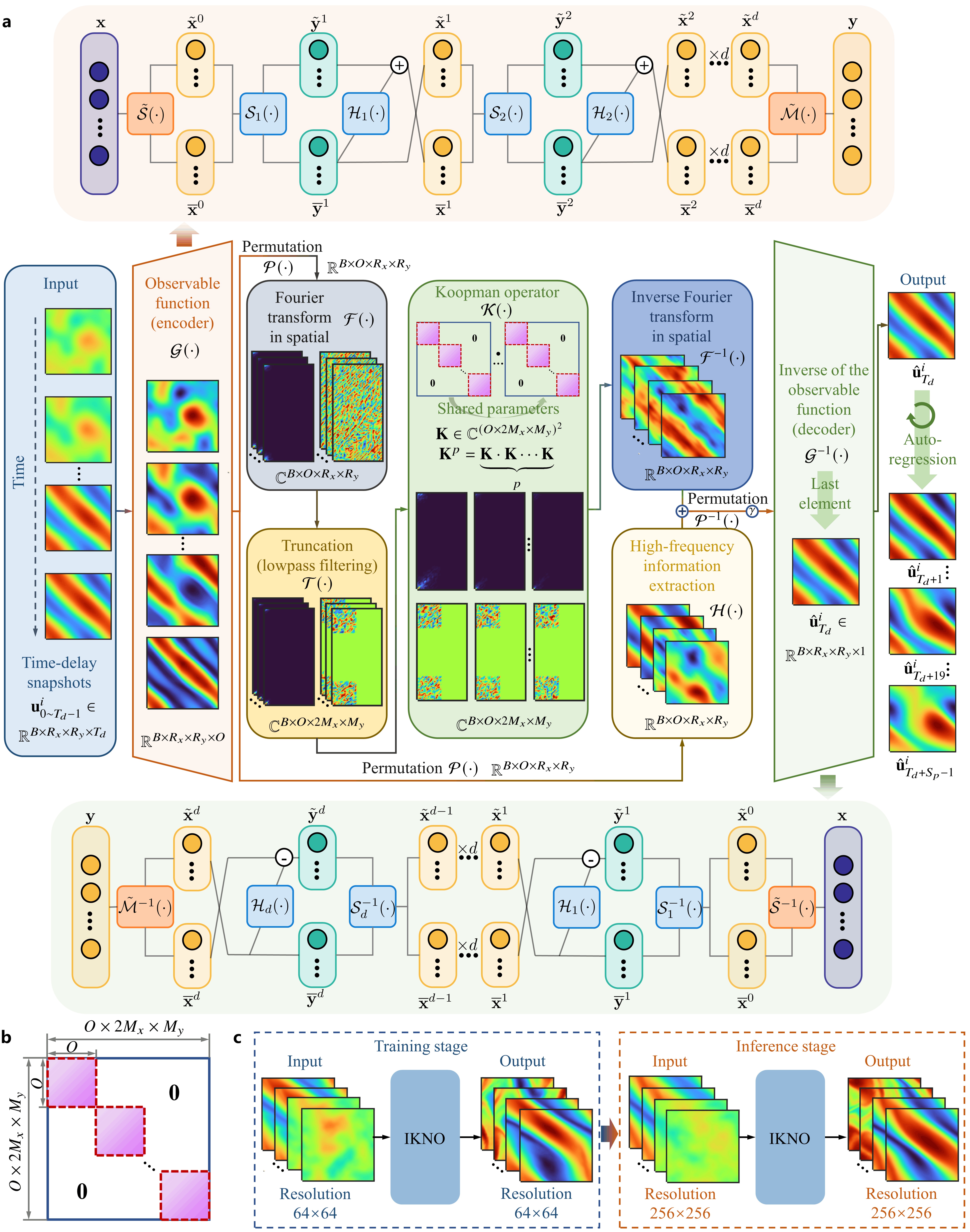}
      \caption{Architecture of the proposed Invertible Koopman Neural Operator (IKNO). \textbf{a} Core components of the IKNO. \textbf{b} Structured linear matrix for approximating the Koopman operator. \textbf{c} The proposed IKNO is resolution-invariant, i.e., the IKNO trained at low resolution can be directly used to predict super-resolution results.}
      \label{fig:ikno_structure}
\end{figure}

\textbf{(1) Nonlinear observable function $\mathcal{G}$}. This part takes $\mathbf{u}^{i}_{0 \thicksim T_{d}-1} \in \mathbb{R}^{B \times R_{1} \times R_{2} \times T_{d}}$ as input. Inspired by the Koopman operator theory, we first transform the input to the observable via $\mathcal{G}(\mathbf{u}^{i}_{0 \thicksim T_{d}-1}) \in \mathbb{R}^{B \times R_{1} \times R_{2} \times O}$, where $O$ (typically greater than $T_{d}$) is the observable's dimension. Note that as we mentioned in \textbf{Subsection 2.3}, it is generally necessary to construct both the observation function and its corresponding inverse, so we innovate here with an INN's forward process to instantiate $\mathcal{G}$, i.e., $O = 2c_{d}$. More specific advantages of this approach will be given in a more detailed statement later. The form of $\mathbf{u}^{i}_{0 \thicksim T_{d}-1}$ can be interpreted as the time-delay coordinates of the original PDE, which is consistent with the Koopman operator's time-delay version. Besides, in the neural operator, $\mathcal{G}$ should be adopted as a spatial-resolution-independent operator to ensure resolution-invariance. Hence, in the concrete implementation, all operations of the INN's forward process equivalent to act on the last dimension of $\mathbf{u}^{i}_{0 \thicksim T_{d}-1}$ only, or alternatively speaking, $\mathcal{G}$ is spatially shared. Otherwise, it leads to unacceptable parameter sizes. This paradigm introduces a potential limitation because the observables do not mix spatial information. However, the Fourier transform can solve this problem.

\textbf{(2) Fourier transform $\mathcal{F}$}. This part takes the permutated output of $\mathcal{G}$ with dimension $\mathbb{R}^{B \times O \times R_{1} \times R_{2}}$ as input. We use the Fourier transform to map the observable measurements to the frequency space, and all subsequent operations are performed on it. The advantage of such processing is the mixing of spatial information in the observables, which nicely compensates for the drawback that $\mathcal{G}$ is spatially shared. Moreover, for a uniform mesh on the Cartesian domain, the Fourier transform can be computed efficiently by the $n$D Fast Fourier Transform (FFT) (in this example, it is 2D FFT on the last two dimensions). Therefore, the computational efficiency is guaranteed. The output of $\mathcal{F}$ is $\mathbb{C}^{B \times O \times R_{1} \times R_{2}}$, which is complex-valued. 

\textbf{(3) Truncation (lowpass filtering) $\mathcal{T}$}. This part takes the output of $\mathcal{F}$ as input. A lowpass filter truncates the observables in the frequency space to remove high-frequency modes, which is motivated by the following physical intuition: the evolution of low-frequency modes over time is usually more stable and gradual, and thus, it is easier to obtain a finite-dimensional approximation of the corresponding Koopman operator. Lowpass filtering is straightforward to implement in frequency space by setting the elements at the specified positions to zero. Here, we denote the cutoff frequencies in the two spatial dimensions as $M_{1}$ and $M_{2}$, respectively. Furthermore, in order to obtain a real signal after the subsequent Fourier inverse transform, the observables should be constrained to be a one-sided Hermitian signal in the Fourier domain. Hence, the lowpass filtering is performed up to the Nyquist frequency in the last dimension, i.e., the output's dimension of $\mathcal{T}$ is $\mathbb{C}^{B \times O \times 2M_{1} \times M_{2}}$. Further, based on this, it can be seen that the introduction of lowpass filtering also helps the IKNO to achieve resolution-invariance, since for the input with arbitrary resolution, lowpass filtering extracts only low-frequency modes with a fixed cutoff frequency under the frequency space, i.e., for the input with arbitrary resolution, the output dimension of $\mathcal{T}$ is constant.

\textbf{(4) Koopman operator approximation $\mathcal{K}$}. This part takes the output of $\mathcal{T}$ with dimension $\mathbb{C}^{B \times O \times 2M_{1} \times M_{2}}$ as input. We obtain the finite-dimensional approximation of the Koopman operator via a linear matrix $\mathbf{K}$ for learning the evolution of the observables' low-frequency modes over time. Since it is under the Fourier domain, $\mathbf{K}$ is a complex-valued matrix with dimension $\mathbb{C}^{(O \times 2M_{1} \times M_{2})^2}$, which is equivalent to treating the flattened input as the observable, with dimension $\mathbb{C}^{B \times (O \times 2M_{1} \times M_{2})}$. Additionally, an adjustable hyperparameter $p \in \mathbb{Z}$ is introduced, allowing us to customize the actual time interval at which the Koopman operator acts. Specifically, we act on the $p$-th power of the linear matrix to the flattened input, i.e., the actual time interval where the Koopman operator acts in the Fourier domain is $\Delta t / p$. This trick does not introduce additional parameters but adds computational cost. Hence, $p$ should be set to a proper value. In the following content, unless otherwise stated, we take $p = 2$. Also, it is clear that if $\mathbf{K}$ is dense, it leads to unacceptable parameter sizes. Hence, we adopt a structured linear matrix to approximate the Koopman operator. As shown in Fig.\ref{fig:ikno_structure}b, the structured $\mathbf{K}$ has is block-diagonal, where each block's dimension is $O \times O$. In this case, the number of trainable parameters of $\mathbf{K}$ is $O \times O \times 2M_{1} \times M_{2}$, similar to a convolutional layer with non-shared weights. Therefore, in the concrete implementation, $\mathbf{K}$ can be directly parameterized as a fourth-order tensor and efficiently computed by Einstein summation.

\textbf{(5) Inverse Fourier transform $\mathcal{F}^{-1}$}. This part takes the output of $\mathcal{K}$ with dimension $\mathbb{C}^{B \times O \times 2M_{1} \times M_{2}}$ as input, and we transform the Koopman operator's prediction back to the original observable space. This procedure can be achieved by the Inverse Fast Fourier Transform (IFFT). It should be noted, however, that the result obtained here does not contain high-frequency modes due to the low-pass filtering.

\textbf{(6) High-frequency information extraction $\mathcal{H}$ and mixing}. This part takes the permutated output of $\mathcal{G}$ with dimension $\mathbb{R}^{B \times O \times R_{1} \times R_{2}}$ as input. This part aims to supplement the high-frequency information that is missing due to the low-pass filtering. The result of Park and Kim \cite{park2022visiontransformerswork} shows that the convolutional layer can extract high-frequency information. Hence, we adopt a convolutional layer with kernel size $1^{2}$ to realize $\mathcal{H}$. The output of $\mathcal{H}$ is $\mathbb{R}^{B \times O \times R_{1} \times R_{2}}$. Then, we mix the high-frequency information with $\mathcal{F}^{-1}$'s output by a point-wise summation. Finally, the activation function, $\gamma$, acts on the mixing result, introducing the necessary complexity for the IKNO. 

\textbf{(7) Inverse of the observable function $\mathcal{G}^{-1}$}. The output of $\gamma$ is considered as a prediction of the observable at the next time step, consistent with the Koopman operator theory. Subsequently, we need to map the observable back to the original physical space using the inverse of the observable function. Note that since the observable function is constructed via an INN's forward process, we we can directly obtain $\mathcal{G}^{-1}$ directly via its corresponding inverse process. This approach does not introduce any additional parameters, does not require additional tasks to the training, and very naturally fulfills the requirement for reconstruction, which is one of the key points of the IKNO.

The final output of $\mathcal{G}^{-1}$ is the prediction result at the next time step with dimension $\mathbb{R}^{B \times R_{1} \times R_{2} \times T_{d}}$. However, due to the utilization of the time-delay coordinates, the output contains redundant information and only the last element is useful, namely $\hat{\mathbf{u}}_{T_{d}}^{i}$. To summarize, the output of IKNO with one iteration can be represented as
\begin{equation}
      \hat{\mathbf{u}}_{T_{d}}^{i} = \mathcal{G}^{-1} \circ \gamma \circ \mathcal{P}^{-1} \circ ( \mathcal{F}^{-1} \circ \mathcal{K} \circ \mathcal{T} \circ \mathcal{F} \circ \mathcal{P} \circ \mathcal{G}(\mathbf{u}^{i}_{0 \thicksim T_{d}-1}) + \mathcal{H} \circ \mathcal{P} \circ \mathcal{G}(\mathbf{u}^{i}_{0 \thicksim T_{d}-1}))(..., T_{d})
      \label{eq:ikno_output}
\end{equation}
Then, as shown in Fig.\ref{fig:ikno_structure}a, iterating the above process using autoregression, we can obtain predictions for multiple (following the problem formulation, $T_{p}$) future time steps, denoted as $\hat{\mathbf{u}}_{T_{d}+1}^{i}, \cdots, \hat{\mathbf{u}}_{T_{d}+T_{p}-1}^{i}$. Moreover, by stacking the above components, we can obtain the IKNO with $l$ numbers of iterations. As shown in Fig.\ref{fig:ikno_structure}c, like other popular neural operators, the proposed IKNO is resolution-invariant, i.e., the IKNO trained at low resolution can be directly used to predict super-resolution results.

\subsection{Mathematical correspondence of each IKNO component in the neural operator theory}

This subsection provides the mathematical correspondence of each IKNO component in the neural operator theory, essential for presenting the rationality of the IKNO's design. Following the general form of the neural operator, $\mathcal{G}$ and $\mathcal{G}^{-1}$ in the proposed IKNO are actually the encoder and decoder, respectively, given by
\begin{equation}
      \hat{\mathcal{J}}: a \mapsto \underbrace{\mathcal{E}(a)}_{\mathcal{G}(a)} = v_{0} \mapsto v_{1} \cdots \mapsto v_{l} \mapsto \underbrace{\mathcal{D}(v_{l})}_{\mathcal{G}^{-1}(v_{l})} = \hat{w}
      \label{eq:neural_operator_ikno}
\end{equation}
The difference is that the IKNO is inspired by the Koopman operator theory so that the encoder and decoder have a more explicit meaning than just the lifting and descending operations: $\mathcal{G}$ is the observable function, while $\mathcal{G}^{-1}$ corresponds to its inverse. The two should satisfy $x = \mathcal{G}^{-1} \circ \mathcal{G}(x)$. In the IKNO, we construct both $\mathcal{G}$ and $\mathcal{G}^{-1}$ with an INN's forward and inverse process instead of parameterizing them separately with two independent shallow MLPs, as in the KNO, and then additionally introducing a reconstruction loss to constrain the reconstruction process during the training procedure. Then, consider the kernel integral operator in $\mathcal{D}_{i, i-1}$, and assume the kernel function has the following form
\begin{equation}
      k(x, y) = \underbrace{(m * m * \cdots * m)}_{p}(x-y)
      \label{eq:kernel_function_ikno}
\end{equation}
where $m: \mathbb{R}^{d_{x}} \mapsto \mathbb{R}^{d_{v} \times d_{v}}$ is a periodic function, and $*$ represents the convolution operation. Based on which, the kernel integral operator can be expressed as the following convolutional form
\begin{equation}
      \int_{\Omega} k(x, y) v_{i-1}(y) dy = \int_{\Omega} (m * m * \cdots * m)(x-y) v_{i-1}(y) dy = (m * m * \cdots * m * v_{i-1})(x)
      \label{eq:kernel_integral_operator_ikno}
\end{equation}
Then, simultaneous Fourier transform and inverse Fourier transform at the right end of Eq.\eqref{eq:kernel_integral_operator_ikno} and applying the convolution theorem yields
\begin{equation}
      \mathcal{F}^{-1} \circ \mathcal{F} ((m * m * \cdots * m * v_{i-1})(x)) = \mathcal{F}^{-1} (\mathcal{F}(m)(\omega) \bullet \cdots \bullet \mathcal{F}(m)(\omega) \bullet \mathcal{F}(v_{i-1})(\omega))(x)
      \label{eq:kernel_integral_operator_fourier_ikno}
\end{equation}
Note that $m$ is periodic, hence the frequency mode $\omega \in \mathbb{Z}^{d_{v}}$. Moreover, $\omega$ can be truncated at the maximal number of modes $M_{i}, i = 1, \cdots, d_{x}$ (lowpass filtering), leading to a finite-dimensional parameterization of $\mathcal{F}(m)$. Hence, $\mathcal{F}(m)$ can be parameterized as a $d_{x} + 2$-order tensor with dimension $\mathbb{C}^{2M_{1} \times \cdots \times M_{d_{x}} \times d_{v} \times d_{v}}$, imposing the conjugate symmetry to ensure the real-valued output of $\mathcal{F}^{-1}$. Then, Eq.\eqref{eq:kernel_integral_operator} can be rewritten as
\begin{equation}
      v_{i} (x) = \mathcal{D}_{i, i-1}(v_{i-1})(x) = \gamma(\mathcal{F}^{-1} (\underbrace{\mathcal{F}(m)(\omega) \bullet \cdots \bullet \mathcal{F}(m)(\omega) \bullet \mathcal{F}(v_{i-1})(\omega))}_{\mathcal{K}}(x) + \mathcal{H}(v)(x))
      \label{eq:kernel_integral_operator_ikno_2}
\end{equation}
where the parameterized $\mathcal{F}(m)$ is the structured linear matrix $\mathbf{K}$, i.e., the Koopman operator $\mathcal{K}$ in the IKNO. The Fourier transform, the inverse Fourier transform, and the activation function in Eq.\eqref{eq:kernel_integral_operator_ikno_2} correspond to each other in the proposed IKNO, while the part of the high-frequency information extraction corresponds to the bias term in the neural operator. At this point, all parts in the proposed IKNO, except for the trivial permutation operation ($\mathcal{P}$ and $\mathcal{P}^{-1}$), have a precise mathematical correspondence with the neural operator theory, ensuring a theoretical basis for the design of the IKNO architecture. 

\subsection{Loss function and training settings}

Consider the training dataset divided in mini-batches (batchsize is $B$) $\mathcal{D} = \{ \mathbf{u}^{i}_{0 \thicksim T_{d}-1}, \mathbf{u}^{i}_{T_{d} \thicksim T_{d}+T_{p}-1} \}_{i = 1}^{N_{B}}$, where $N_{B}$ is the number of mini-batches, and $\mathbf{u}^{i}_{T_{d} \thicksim T_{d}+T_{p}-1} = \{ \mathbf{u}_{T_{d}+1}^{i}, \cdots, \mathbf{u}_{T_{d}+T_{p}-1}^{i} \}$. The prediction result of the IKNO is denoted as $\{ \hat{\mathbf{u}}_{T_{d}+1}^{i}, \cdots, \hat{\mathbf{u}}_{T_{d}+T_{p}-1}^{i} \}$. The loss function to train the IKNO is defined as the relative $L_{2}$ error, given by
\begin{equation}
      \mathcal{L}_{pred} = \frac{1}{T_{p}} \sum_{j = T_{d}}^{T_{d}+T_{p}-1} \frac{\| \hat{\mathbf{u}}_{j}^{i} - \mathbf{u}_{j}^{i} \|_{2}}{\| \mathbf{u}_{t}^{i} \|_{2}}
      \label{eq:loss_function}
\end{equation}
where $\| \cdot \|_{2}$ represents the $L_{2}$ norm. The training process is implemented by the Adam optimizer \cite{kingma2017adam} with a initial learning rate of $10^{-3}$, and the batch size is set to 10. Unless otherwise emphasized, the training epoches is set to 500, and the learning rate is havled every 100 epoches.

\textbf{Remark 1}. As emphasized in the previous section, the proposed IKNO constructs the observable function and its corresponding inverse via the INN. With arbitrary INN parameters, it is natural to satisfy $x = \mathcal{G}^{-1} \circ \mathcal{G}(x)$. Therefore, introducing an additional reconstruction loss in the IKNO's training process is unnecessary, which is an important optimization task in the original KNO \cite{XIONG2024113194, xiongKoopmanLabMachineLearning2023, CAO2024544}. This property is undoubtedly beneficial because there is only one loss function during the training process, and there is no need to balance the weights between different losses.

\section{Numerical and real-world examples}

This section fully illustrates the effectiveness of our proposed IKNO through rich numerical and real-world PDE examples, including the prediction performance and the highlighted zero-shot super-resolution prediction results benefiting from the resolution-invariance. Moreover, in each example, the IKNO is compared in detail with the FNO \cite{li2021fourierneuraloperatorparametric} and KNO \cite{XIONG2024113194,xiongKoopmanLabMachineLearning2023} to demonstrate its superiority. For a fair comparison, all methods are trained and tested under the same conditions. The maximal number of modes $M_{i}, i = 1, \cdots, d_{x}$ is set to 16, and $d_{v}$ is set to 32. The number of iterations ($l$) is set to 4 for the IKNO, FNO, and 8 for the KNO. 

\subsection{1D Burgers equation}

As the first example, we consider the 1D Burgers equation, which is a classic nonlinear PDE, defined by
\begin{equation}
      \begin{aligned}
            \partial_{t} u(x, t) + u(x, t) \partial_{x} u(x, t) &= \nu \partial_{xx}u(x, t), x \in [0, 1], t \in (0, 1] \\
            u(x, 0) &= u_{0}(x), x \in [0, 1]
      \end{aligned}
      \label{eq:burgers}
\end{equation}
where$u(x, t)$ represents the velocity field, $\nu \in \mathbb{R}^{+}$ is the viscosity, and here we set $\nu = 0.01$. The initial condition is randomly sampled from a given Gaussian random field with a Riesz kernel, where a degree of correlation is introduced for neighboring mesh points \cite{Lu_2021}. Eq.\eqref{eq:burgers} is solved based on the split method and the forward Euler method on a uniform spatial mesh with 1024 resolution. The time advancement interval, i.e., $\Delta t$, is 0.1s. The training set contains 1800 samples downsampled to the resolution of $2^{5} = 32$. Then, the test set contains 200 samples with a resolution from $2^{5}$ to $2^{10}$ for evaluating the proposed method's basic and zero-shot super-resolution prediction results, respectively. The objective is to learn a neural operator based on the IKNO that maps the velocity field at the initial $T_{d} = 10$ time steps to the future velocity field at the next $T_{p} = 90$ time steps, i.e., $u|_{[0, 1] \times \{ t_{0}-9\Delta t, \cdots, t_{0} \}} \mapsto u|_{[0, 1] \times \{ t_{0} + \Delta t, \cdots, t_{0} + 90 \Delta t \}}$.

\begin{figure}[htbp]
      \centering 
      \includegraphics[width = 0.99\textwidth]{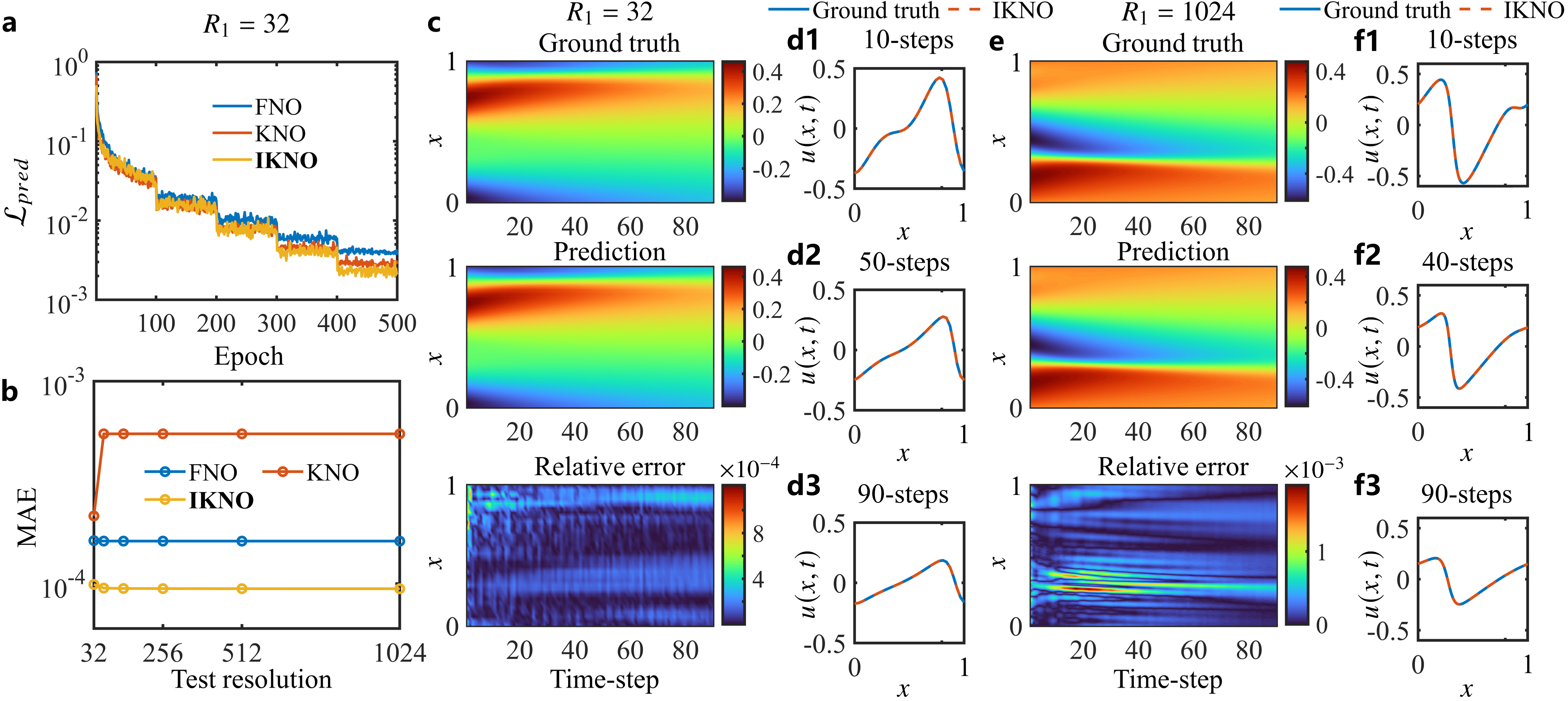}
      \caption{Results of the 1D Burgers equation. \textbf{a} Comparison of the loss function. The proposed IKNO shows better convergence during training, achieving the lowest loss. Note that the training process is performed at the resolution of 32. \textbf{b} Comparison of the resolution-invariance and super-resolution prediction performance for different methods. The proposed IKNO has the minimum MAE at all resolutions. \textbf{c} The ground truth, prediction, and relative error distribution of the IKNO at the resolution of 32. Three temporal snapshots are also portrayed in \textbf{d1}-\textbf{d3}. \textbf{e} The ground truth, prediction, and error distribution of the IKNO at the resolution of 1024, 32$\times$ the training set. Three temporal snapshots are also portrayed in \textbf{f1}-\textbf{f3}. Even under the zero-shot super-resolution condition, the proposed IKNO still captures sharp jumps and discontinuity in the velocity field with high accuracy.}
      \label{fig:burgers}
\end{figure}

The relevant results of the 1D Burgers equation are shown in Fig.\ref{fig:burgers}. Firstly, Fig.\ref{fig:burgers} compares the loss function during training for the listed different methods. It can be observed that the KNO and IKNO show better convergence during training (than the FNO), while our proposed IKNO achieves the lowest loss. A plausible explanation for this is that the INN introduced in the IKN implements a parameter-independent construction of the observation function and its inverse; its unique structure ensures that $x = \mathcal{G}^{-1} \circ \mathcal{G}(x)$ holds strictly even under arbitrary INN's parameters. Hence, the reconstruction task constrained by the reconstruction loss is accomplished naturally, a necessary optimization task in the original KNO. On the one hand, we don't need to tediously try to balance the weights between the prediction and reconstruction loss; on the other hand, this directly avoids potential gradient direction conflict, which is very common in the multi-task learning process. This advantage leads to more accurate prediction results. As can be seen from Fig.\ref{fig:burgers}b, all three discussed methods exhibit great resolution-invariance, i.e., the Mean Absolute Error (MAE) remains essentially constant as the resolution of the test set increases; note that the training process is performed at the resolution of 32. Among them, the proposed IKNO has the best accuracy at all resolutions. Visually, as illustrated in Fig.\ref{fig:burgers}c and d1-d3, there is an excellent match between the prediction and ground truth with errors on the order of $10^{-4}$. Moreover, the test result at the resolution of 1024 ($32 \times$ the training set) is shown in Fig.\ref{fig:burgers}e, and significant agreement between the prediction and ground truth can still be observed, demonstrating the IKNO's ability for zero-shot super-resolution prediction. For a more detailed assessment, three temporal snapshots (10-, 40-, and 90-steps) are also portrayed in Fig.\ref{fig:burgers}f1-f3, showing that the proposed IKNO captures sharp jumps and discontinuity in the velocity field caused by low viscosity with high accuracy. The above results illustrate the effectiveness and superiority of our proposed IKNO.

\subsection{2D shallow water equation}

As the second example, the 2D shallow water equation is discussed, which is utilized to describe free-surface flows in large-scale oceanic and atmospheric dynamics, defined by
\begin{equation}
      \begin{aligned}
            \partial_{t} h + \nabla \cdot (h \mathbf{v}) &= 0 \\
            \partial_{t} h \mathbf{v} + \nabla \cdot (h \mathbf{v} \otimes \mathbf{v}) + g h \nabla h &= 0
      \end{aligned}
      \quad (x, y) \in [-2.5, 2.5]^{2}, t \in (0, 1]
      \label{eq:shallow_water}
\end{equation}
where $h$ is the water depth and $\mathbf{v} = [u, v]$ describes the velocity field (in the horizontal and vertical directions). For simplicity, $h(x, y, t)$ is abbreviated as $h$, and $\mathbf{v}$ is the same. $g$ represents the gravitational acceleration. Then, 2D radial dam break problems are considered, i.e., the initial condition is given as
\begin{equation}
      h(x, y, 0) = \begin{cases}
            2.0, \sqrt{x^{2} + y^{2}} \leq r \\
            1.0, \sqrt{x^{2} + y^{2}} > r
      \end{cases}
      \label{eq:shallow_water_initial}
\end{equation}
where $r$ is the radius, randomly sampled from a uniform distribution in the range of $[0.3, 0.7]$. With $\Delta t = 0.01$s, the training set contains 1800 samples downsampled to the resolution of $64 \times 64$. The test set contains 200 samples with a resolution from $64 \times 64$ to $128 \times 128$ for evaluating the proposed method's basic and zero-shot super-resolution prediction results, respectively. More details about the data set are provided in reference \cite{takamoto2024pdebenchextensivebenchmarkscientific}. The objective is to learn a neural operator based on the IKNO that maps the water depth at the initial $T_{d} = 10$ time steps to the future water depth at the next $T_{p} = 90$ time steps, i.e., $h|_{[-2.5, 2.5]^{2} \times \{ t_{0}-9\Delta t, \cdots, t_{0} \}} \mapsto h|_{[-2.5, 2.5]^{2} \times \{ t_{0} + \Delta t, \cdots, t_{0} + 90 \Delta t \}}$. 

\begin{figure}[htbp]
      \centering
      \includegraphics[width = 0.99\textwidth]{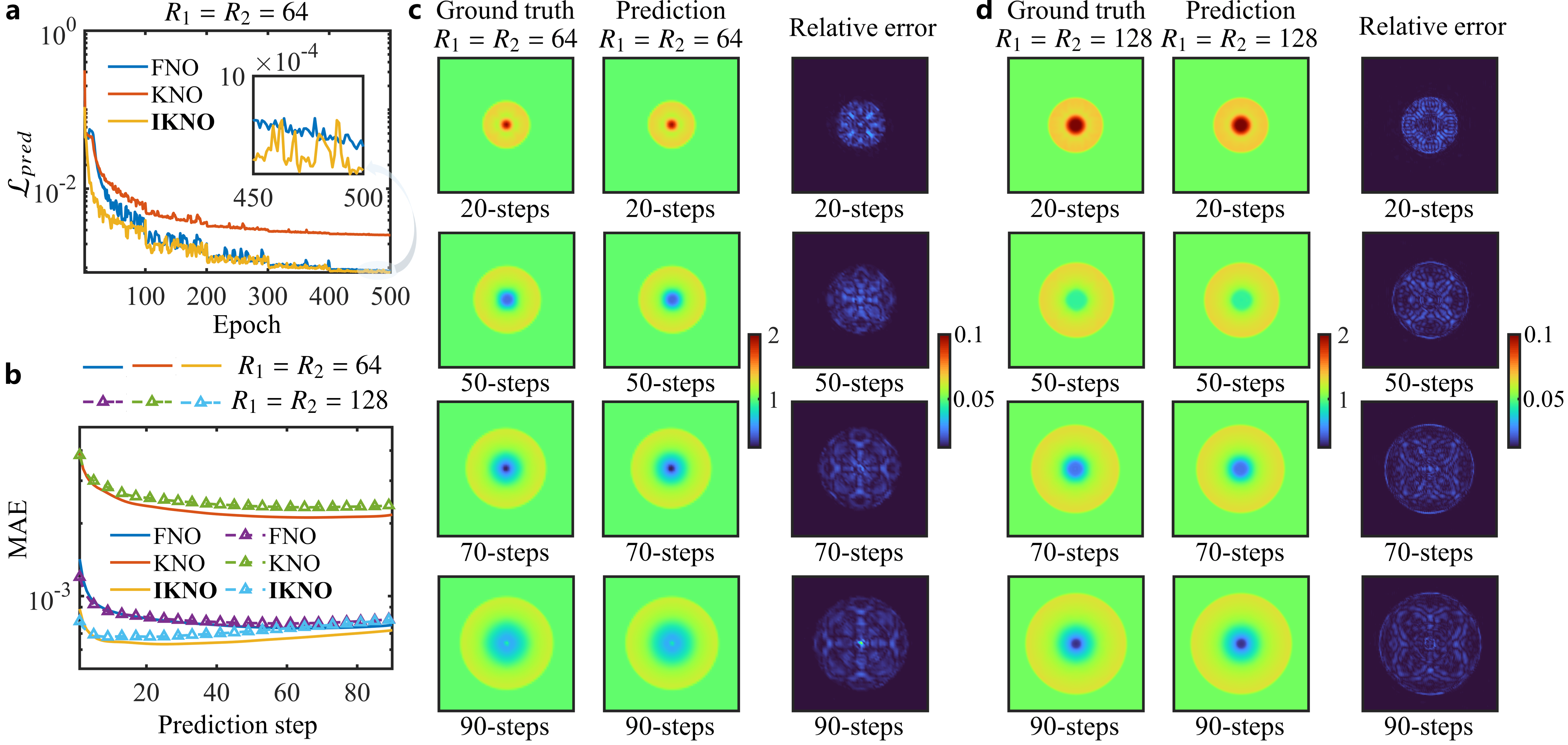}
      \caption{Results of the 2D shallow water equation. \textbf{a} Comparison of the loss function. The proposed IKNO shows better convergence during training, achieving the lowest loss. Note that the training process is performed at the resolution of $64 \times 64$. \textbf{b} Comparison of the MAE with different prediction steps. Results of resolution-invariance are also plotted, where the methods are evaluated at the resolution of $64 \times 64$ and $128 \times 128$ (2$\times$ the training set). \textbf{c} The ground truth, prediction, and relative error distribution of the IKNO at the resolution of $64 \times 64$. \textbf{e} The ground truth, prediction, and error distribution of the IKNO at the resolution of $128 \times 128$, 2$\times$ the training set.}
      \label{fig:shallow_water}
\end{figure}

The corresponding results are summarized in Fig.\ref{fig:shallow_water}. Fig.\ref{fig:shallow_water}a presents that all three discussed methods show satisfactory convergence, with the relative $L_{2}$ error at the end of training decreasing to the order of $10^{-3}$. Among these, FNO and our proposed IKNO are more competitive candidates. The variation of MAE with the prediction step on the test set is illustrated in Fig.\ref{fig:shallow_water}b. It can be observed that the proposed IKNO has more minor errors in both the standard and super-resolution cases. Furthermore, it can also be seen that the listed methods have great resolution-invariance, i.e., the error hardly changes significantly with the resolution size. More intuitive results are illustrated in Fig.\ref{fig:shallow_water}c and d. The IKNO achieves accurate long-term predictions (90 time-steps), where only the water depth in the initial 10 time steps is provided. Then, an interesting pattern shown in Fig.\ref{fig:shallow_water}c and d is that the errors are concentrated within the shock front region. A persuasive explanation for this is that in a dam break scenario, the water depth abruptly changes at the shock front, while in other areas where the water wave has not yet spread, the water depth remains constant. The proposed IKNO accurately captures these characteristic dynamics, leading to reliable predictions over long-term horizons, even under the zero-shot super-resolution condition (shown in Fig.\ref{fig:shallow_water}d), demonstrating its effectiveness. 

\subsection{2D incompressible Navier-Stokes equation}

Then, we consider a more challenging task, the 2D incompressible Navier-Stokes equation, which is a fundamental model in fluid dynamics defined by
\begin{equation}
      \begin{aligned}
            \partial_t \omega + \mathbf{u} \cdot \nabla \omega &= \nu \nabla^2 \omega + f \\
            \nabla \cdot \mathbf{u} &= 0 \\
            \omega &= \nabla \times \mathbf{u}
      \end{aligned}
      \quad (x, y) \in [0, 1]^{2}, t \in (0, T]
      \label{eq:navier_stokes}
\end{equation}
where $\mathbf{u}$ is the velocity field, $\omega$ is the vorticity, $\nu \in \mathbb{R}^{+}$ is the viscosity, and $f$ represents the external force, given as
\begin{equation}
      f(x, y) = 0.1 ( \sin (2 \pi (x + y)) + \cos (2 \pi (x + y)) )
\end{equation}
A Gaussian random field with a Riesz kernel generates the initial vorticity. Eq.\eqref{eq:navier_stokes} is solved based on the stream-function formulation with a pseudospectral method under the discretization mesh of $256 \times 256$. Then, consider the vorticities $\nu = 10^{-3}, 10^{-4}$ to establish the dataset. The training set contains 1000 and 8000 samples downsampled to $64 \times 64$ separately. The test set in both cases contains 200 samples. Additionally, to evaluate the resolution-invariance, another test set with $\nu = 10^{-4}$ contains 20 samples with original $256 \times 256$ resolution is also constructed. Since the velocity field can be easily obtained through the vorticity field, we aim to learn a neural operator that maps the vorticity at the initial $T_{d} = 10$ time steps to the future vorticity. For $\nu = 10^{-3}$, $T_{d} = 40$, i.e., $\omega|_{[0, 1]^{2} \times \{ t_{0}-9\Delta t, \cdots, t_{0} \}} \mapsto \omega|_{[0, 1]^{2} \times \{ t_{0} + \Delta t, \cdots, t_{0} + 40 \Delta t \}}$; for $\nu = 10^{-4}$, $T_{d} = 20$, i.e., $\omega|_{[0, 1]^{2} \times \{ t_{0}-9\Delta t, \cdots, t_{0} \}} \mapsto \omega|_{[0, 1]^{2} \times \{ t_{0} + \Delta t, \cdots, t_{0} + 20 \Delta t \}}$.

\begin{figure}[htbp]
      \centering
      \includegraphics[width = 0.99\textwidth]{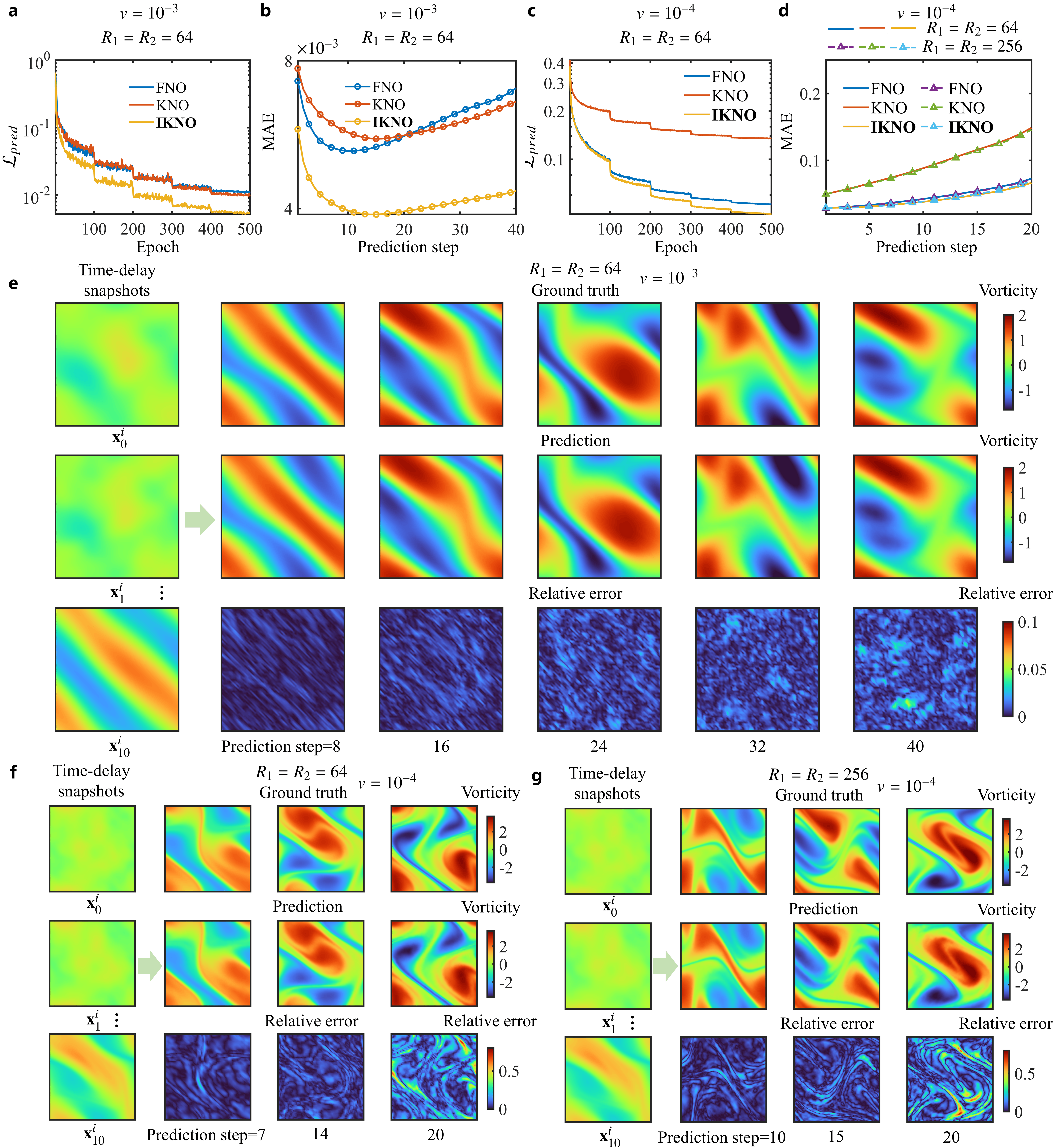}
      \caption{Results of the 2D incompressible Navier-Stokes equation. \textbf{a} Comparison of the loss function with viscosity $\nu = 10^{-3}$. The proposed IKNO shows better convergence during training, achieving the lowest loss. \textbf{b} Comparison of the MAE under different prediction steps with viscosity $\nu = 10^{-3}$. \textbf{c} Comparison of the loss function with viscosity $\nu = 10^{-4}$. Note that the training process is performed at the resolution of $64 \times 64$. \textbf{d} Comparison of the MAE under different prediction steps with viscosity $\nu = 10^{-4}$. Results of resolution-invariance are also plotted, where the methods are evaluated at the resolution of $64 \times 64$ and $256 \times 256$ (4$\times$ the training set). \textbf{e} The ground truth, prediction, and relative error distribution of the velocity obtained by the IKNO at the resolution of $64 \times 64$, with $\nu = 10^{-3}$. \textbf{f} and \textbf{g} The ground truth, prediction, and relative error distribution of the velocity obtained by the IKNO at the resolution of $64 \times 64$ and $256 \times 256$, respectively, with $\nu = 10^{-4}$.}
      \label{fig:navier_stokes_equation}
\end{figure}

Fig.\ref{fig:navier_stokes_equation} provides the relevant data-driven modeling contents of the 2D incompressible Navier-Stokes equation. Firstly, from Fig.\ref{fig:navier_stokes_equation}a and c, we can find that the proposed IKNO demonstrates better convergence compared to the FNO and KNO, highlighting its superiority. Moreover, it can also be observed that although the KNO achieves competitive results compared to the FNO in the $\nu = 10^{-3}$ case, its loss function value is significantly higher than that of the FNO and IKNO in the more complex $\nu = 10^{-4}$ case, indicating its potential limitations in handling complex systems. The above claims are further supported by Fig.\ref{fig:navier_stokes_equation}b and d. The proposed IKNO performs best in all test sets, with resolution-invariance and lowest MAE. In contrast, KNO's predictions are close to or even better than FNO's under $\nu = 10^{-3}$ but have significant discrepancies with IKNO's and FNO's in Fig.\ref{fig:navier_stokes_equation}d. More intuitive visual results are plotted in Fig.\ref{fig:navier_stokes_equation}e-g. Due to the small viscosity, Eq.\eqref{eq:navier_stokes} exhibits intricate dynamics, resulting in a highly variable vorticity field in time scale. With the given vorticity field in the initial 10 time steps, our method achieves accurate long-term predictions over 40 time steps (shown in Fig.\ref{fig:shallow_water}e). Then, for a smaller viscosity coefficient ($\nu = 10^{-4}$), this leads to a larger Reynolds number, making the dynamics of Eq.\eqref{eq:navier_stokes} more complex and even chaotic. Under this scenario, the proposed IKNO still captures the vorticity field's evolution with acceptable accuracy over 20 time steps (shown in Fig.\ref{fig:shallow_water}f). Meanwhile, benefiting from the resolution-invariance, the IKNO trained at $64 \times 64$ resolution can be directly used for higher resolution prediction with almost constant accuracy (shown in Fig.\ref{fig:navier_stokes_equation}g). These results indicate that our proposed method is still effective in handling complex systems.

\subsection{2D Darcy flow in a triangular domain with a notch}

This subsection discusses the 2D Darcy flow, defined by
\begin{equation}
      \begin{aligned}
            -\nabla \cdot (\kappa(x, y) \nabla p(x, y)) = f(x, y) \\
      \end{aligned}
      \quad (x, y) \in \bar{ \Omega }
      \label{eq:darcy_flow}
\end{equation}
where $\kappa(x, y)$ is the permeability field, $p(x, y)$ denotes the pressure field, and $f(x, y)$ represents the source term. Here we fix $\kappa(x, y) = 0.1$ and $f(x, y) = -1$. It should be emphasized that, unlike the examples in previous subsections, the solution domain $\bar{\Omega}$ is a triangular region with a notch. Our goal is to learn a neural operator that maps the boundary condition, $p_{bc} (x, y), (x, y) \in \partial \bar{\Omega}$, to the steady pressure field. 

\begin{figure}[htbp]
      \centering
      \includegraphics[width = 0.99\textwidth]{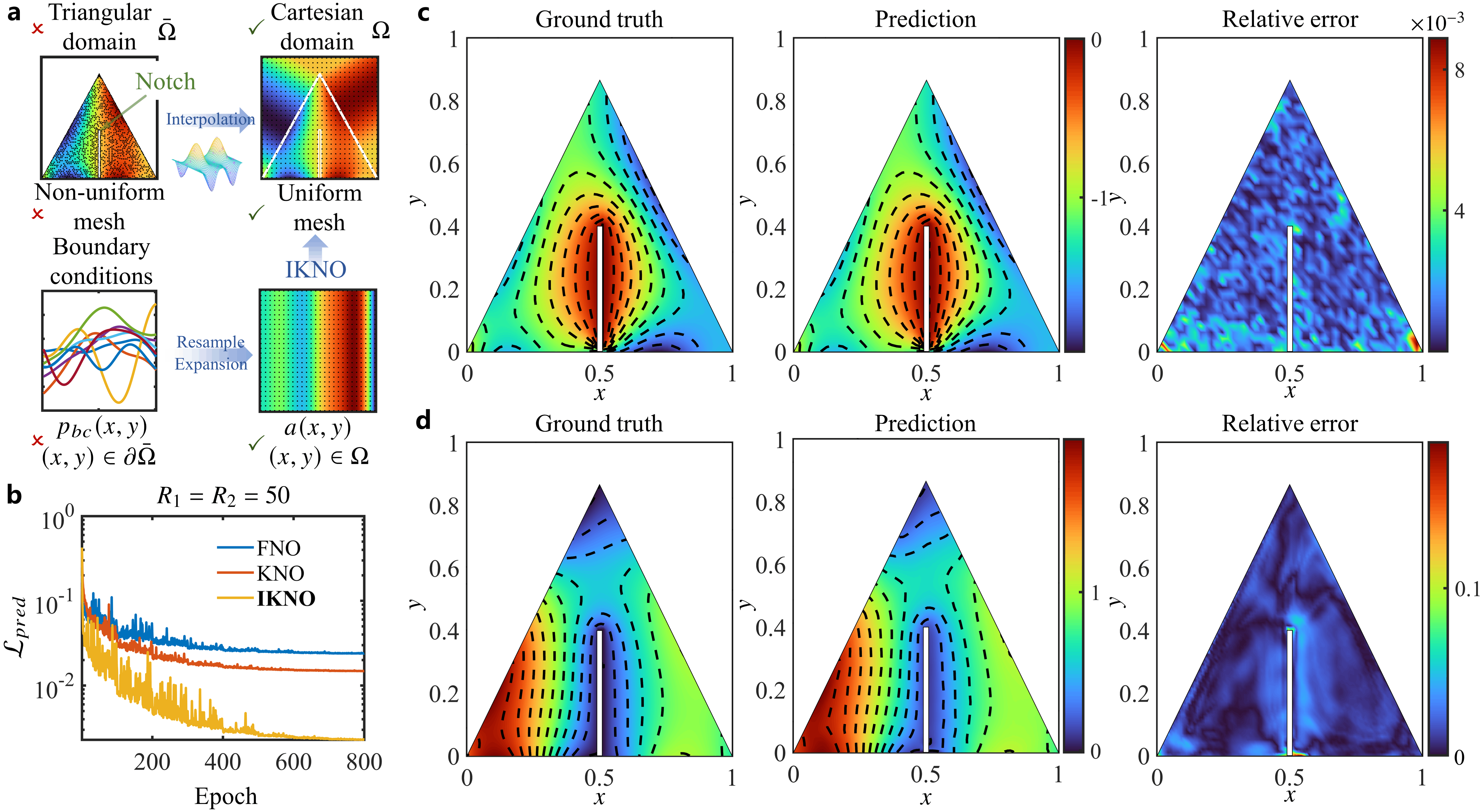}
      \caption{Results of the 2D Darcy flow in a triangular domain with a notch. \textbf{a} A triangular domain with a notch and boundary conditions are preprocessed to allow FFT. \textbf{b} Comparison of the loss function. The proposed IKNO shows better convergence during training, achieving the lowest loss. Note that the training process is performed at the resolution of $50 \times 50$. \textbf{c} The ground truth, prediction, and relative error distribution of the pressure field obtained by the IKNO at the resolution of $50 \times 50$. \textbf{d} The ground truth, prediction, and relative error distribution of the pressure field obtained by the IKNO at the resolution of $100 \times 100$, 2$\times$ the training set.}
      \label{fig:darcy_flow}
\end{figure}

Note that $\bar{ \Omega }$ is not a Cartesian domain, and the domain of definition of $p_{bc} (x, y)$ is also not consistent with the target; hence the proposed IKNO is not directly applicable, the FNO and KNO are also identical. However, inspired by reference \cite{luComprehensiveFairComparison2022}, the listed approaches can be made usable with some simple preprocessing. Specifically, as shown in Fig.\ref{fig:darcy_flow}a, through interpolation, a pressure field defined within a triangular region with a notch can be converted to a solution defined over a Cartesian domain, $\Omega$, even though the solution outside the triangular region is practically meaningless. Subsequently, employing resample and dimension extension, $p_{bc} (x, y)$ defined on the triangular boundary can be changed into a function formally defined on $\Omega$, $a (x, y)$. Based on these, the learning task can be re-formulated as $a (x, y) \mapsto p (x, y), (x, y) \in \Omega$, which is standard and allows FFT. 

A Gaussian process is adopted to generate boundary conditions. With the preprocess shown in Fig.\ref{fig:darcy_flow}a, the training set contains 1900 samples downsampled to the resolution of $50 \times 50$. The test set contains 100 samples with a resolution from $50 \times 50$ and $100 \times 100$ for evaluating the proposed method's basic and zero-shot super-resolution prediction results, respectively. Fig.\ref{fig:darcy_flow}b compares loss function curves of different methods during training; note that more epochs (800) are adopted in this example. We can observe a significant oscillation in the early training stage, indicating a more challenging task. Surprisingly, the KNO gains better results in this example than the FNO, showing its potential advantages in some specific problems. However, our proposed IKNO memorably outperforms them, with losses almost reducing an order magnitude. This result highlights the IKNO's superiority. Then, the ground truth and prediction are presented in Fig.\ref{fig:darcy_flow}c, showing remarkable consistency with errors on the order of $10^{-3}$. Then, the zero-shot super-resolution prediction is set out in Fig.\ref{fig:darcy_flow}d, where the IKNO is trained on the resolution of $50 \times 50$ and evaluated on the resolution of $100 \times 100$. Errors increase somewhat and are mainly concentrated near notch edges. However, in the vast majority of domains, the IKNO's predictions continue to be accurate, showing resolution-invariance. The above results show that the IKNO is still effective in problems with complex geometries defined on non-Cartesian domains, by introducing preprocessing strategies such as interpolation.

\subsection{2D compressible Euler equation for airfoil flow field}

Here, we consider the problem of airfoil flow field prediction, which is controlled by the 2D compressible Euler equation
\begin{equation}
      \begin{aligned}
            \partial_{t} \rho + \nabla \cdot (\rho \mathbf{u}) &= 0 \\
            \partial_{t} (\rho \mathbf{u}) + \nabla \cdot (\rho \mathbf{u} \otimes \mathbf{u} + p \mathbf{I}) &= 0 \\
            \partial_{t} E + \nabla \cdot ((E + p) \mathbf{u}) &= 0
      \end{aligned}
      \quad (x, y) \in \bar{\Omega}, t \in (0, T]
      \label{eq:euler}
\end{equation}
where $\rho$ is the fluid density, $\mathbf{u} = [u, v]$ is the velocity field, $p$ is the pressure field, $E$ is the total energy, and $\mathbf{I}$ represents an identity matrix. $\bar{\Omega}$ is a fluid solve domain around an airfoil, where no-penetration condition is introduced on the airfoil surface. Hence, $\bar{\Omega}$ is a non-Cartesian domain. 

\begin{figure}[htbp]
      \centering
      \includegraphics[width = 0.99\textwidth]{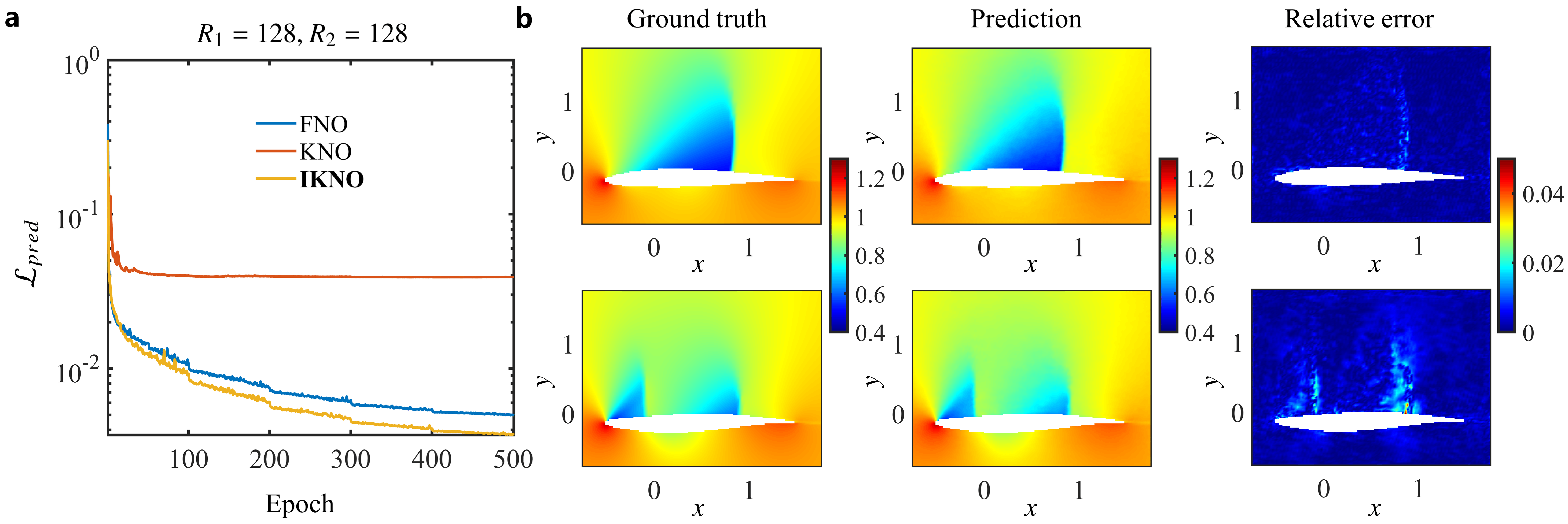}
      \caption{Results of the 2D compressible Euler equation for airfoil flow field, where the interpolation method is introduced to solve the problem of non-Cartesian domain. \textbf{a} Comparison of the loss function. The proposed IKNO shows better convergence during training, achieving the lowest loss. \textbf{b} The ground truth, prediction, and relative error distribution of the density field obtained by the IKNO with different airfoil shapes.}
      \label{fig:airfoils_interpolation}
\end{figure}

The airfoil geometry is parameterized based on the RAE2822 airfoil \cite{LYE2021113575}, i.e., perturbative deformation of the RAE2822 airfoil shape \cite{raonicConvolutionalNeuralOperators2023b}. Steady-state solution of Eq.\eqref{eq:euler} is obtained with the far-field freestream boundary condition, where the distant incoming flow's fluid density $\rho^{\infty} = 1.0$, Mach number $M^{\infty} = 0.729$ (subsonic flow), pressure $p^{\infty} = 1.0$ and angle of attack $\alpha = 2.31^{\circ}$. The goal of the neural operator is to learn the mapping of different airfoil shapes to the steady-state density field. Although $\bar{\Omega}$ is a non-Cartesian domain, by adopting the preprocessing strategy described in the previous subsection (Fig.\ref{eq:darcy_flow}a), the density field can be transformed to data defined on a uniform $128 \times 128$ mesh points in the Cartesian domain, $\Omega \in [-0.75, 1.75]^2$, denoted as $\rho^{steady}(x, y), (x, y) \in \Omega$. Subsequently, the airfoil shape can also be described as a function $ a(x, y)$ that acts on $\Omega$, where the region within the airfoil is set to 1, and the fluid solution domain is taken to be 0. Hence, the learning task is re-formulated as $a(x, y) \mapsto \rho^{steady} (x, y), (x, y) \in \Omega$. The training and test sets contain 800 and 200 samples, respectively. Fig.\ref{fig:airfoils_interpolation} summarizes the relevant results. It can be seen that the proposed IKNO still maintains the best convergence, while the KNO suffers from training difficulties (shown in Fig.\ref{fig:airfoils_interpolation}a). Fig.\ref{fig:airfoils_interpolation}b compares the ground truth and prediction density fields with different airfoil shapes, showing that the proposed IKNO can accurately predict the density field with errors on the order of $10^{-2}$ only based on the information of the airfoil shape, which is not a trivial task.

\begin{figure}[htbp]
      \centering
      \includegraphics[width = 0.99\textwidth]{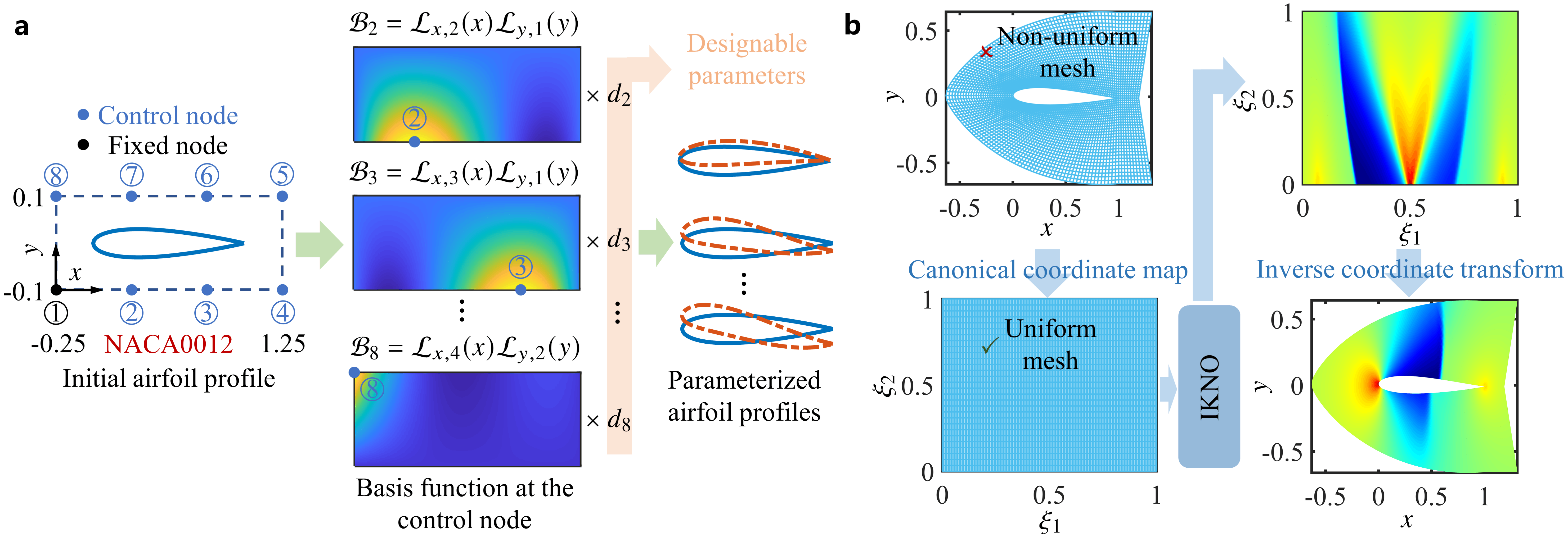}
      \caption{Diagram of airfoil geometry parameterization and canonical coordinate map. \textbf{a} Realization of designable airfoils by lervaging the initial airfoil, basis functions, and their coefficients. \textbf{b} Canonical coordinate maping of non-uniform mesh around the airfoil.}
      \label{fig:airfoils_parameterization_and_geometry}
\end{figure}

Moreover, another scenario is discussed, where the airfoil geometries are parameterized based on the initial airfoil profile (NACA0012) and basis functions, as illustrated in Fig.\ref{fig:airfoils_parameterization_and_geometry}a. Specifically, the dimensionless shape of NACA0012 can be defined as a set of spatial points, $(x_{airfoil}^{NACA}, y_{airfoil}^{NACA}) \in \partial \bar{ \Omega } $, where $\partial \bar{ \Omega}$ denotes the airfoil surface. A new airfoil can be obtained through basis functions at the control nodes defined at a rectangle box, described as 
\begin{equation}
      \begin{aligned}
            x_{airfoil}^{new} &= x_{airfoil}^{NACA} \\
            y_{airfoil}^{new} &= y_{airfoil}^{NACA} + \sum_{i=2}^{4} b_{i1} \mathcal{L}_{x, i}(x_{airfoil}^{NACA}) \mathcal{L}_{y, 1}(x_{airfoil}^{NACA}) + \sum_{j=1}^{4} b_{j2} \mathcal{L}_{x, i}(x_{airfoil}^{NACA}) \mathcal{L}_{y, 2}(x_{airfoil}^{NACA})
      \end{aligned}
      \label{eq:airfoil_parameterization}
\end{equation}
where $\mathcal{L}_{x, i}, i = 1, \cdots, 4$, $\mathcal{L}_{y, j}, j = 1, 2$ are the Lagrange interpolation basis functions, and $b_{i1}, i = 2, \cdots, 4, b_{j2}, j = 1, \cdots, 4$ are the coefficients to control deformation. For ease of presentation, the subscripts of the coefficients are abbreviated as $b_{i}, i = 2, \cdots, 8$. By assigning different values to $d_{i}$, a range of differently shaped airfoil profiles can be obtained, i.e., $d_{i}$ works as designable parameters to control the airfoil shape. 

It should be noted that the problem of the non-Cartesian domain still exists in this scenario. Unlike the treatment of introducing interpolation in previous examples, inspired by literature \cite{JMLR:v24:23-0064}, here we use the canonical coordinate map to solve this issue. As illustrated in Fig.\ref{fig:airfoils_parameterization_and_geometry}b, the airfoil field is solved on $221 \times 51$ non-uniform but structured elliptic mesh. Thus, there exists an explicit canonical coordinate transformation that maps points $(x, y)$ on $\bar{\Omega}$ one-to-one to points defined on a standard Cartesian domain, $(\xi_{1}, \xi_{2}) \in \Omega = [0, 1]^{2}$. Precisely, this canonical coordinate transformation unfolds the structured elliptic mesh radially and circumferentially and then adjusts it to a uniform grid spacing. Based on this, we can act the neural operator under the space of $(\xi_{1}, \xi_{2})$. Subsequently, the obtained results can be transformed into the actual physical space by inverse coordinate transformation, which must exist since the adopted canonical coordinate map is a bijection. Then, the neural operator is utilized to map the airfoil shape to the steady-state pressure and velocity (absolute value) fields simultaneously, resulting in a more challenging task. The inputs are the spatial coordinates of the mesh points in the physical space corresponding to $(\xi_{1}, \xi_{2})$, denoted as $[a_{x}, a_{y}] (\xi_{1}, \xi_{2})$, which contains geometric information of the airfoil. Hence, the learning task is formulated as $[a_{x}, a_{y}] (\xi_{1}, \xi_{2}) \mapsto [p^{steady}, u^{steady}_{abs}] (\xi_{1}, \xi_{2})$. The training and test sets contain 1000 and 200 samples, where $d_{i}$ is randomly sampled from a uniform distribution in the range of $[-0.05, 0.05]$, and the far-field freestream boundary is set as $\rho^{\infty} = 1.0$, $p^{\infty} = 1.0$, $M^{\infty} = 0.8$, and $\alpha = 0^{\circ}$. 

\begin{figure}[htbp]
      \centering
      \includegraphics[width = 0.98\textwidth]{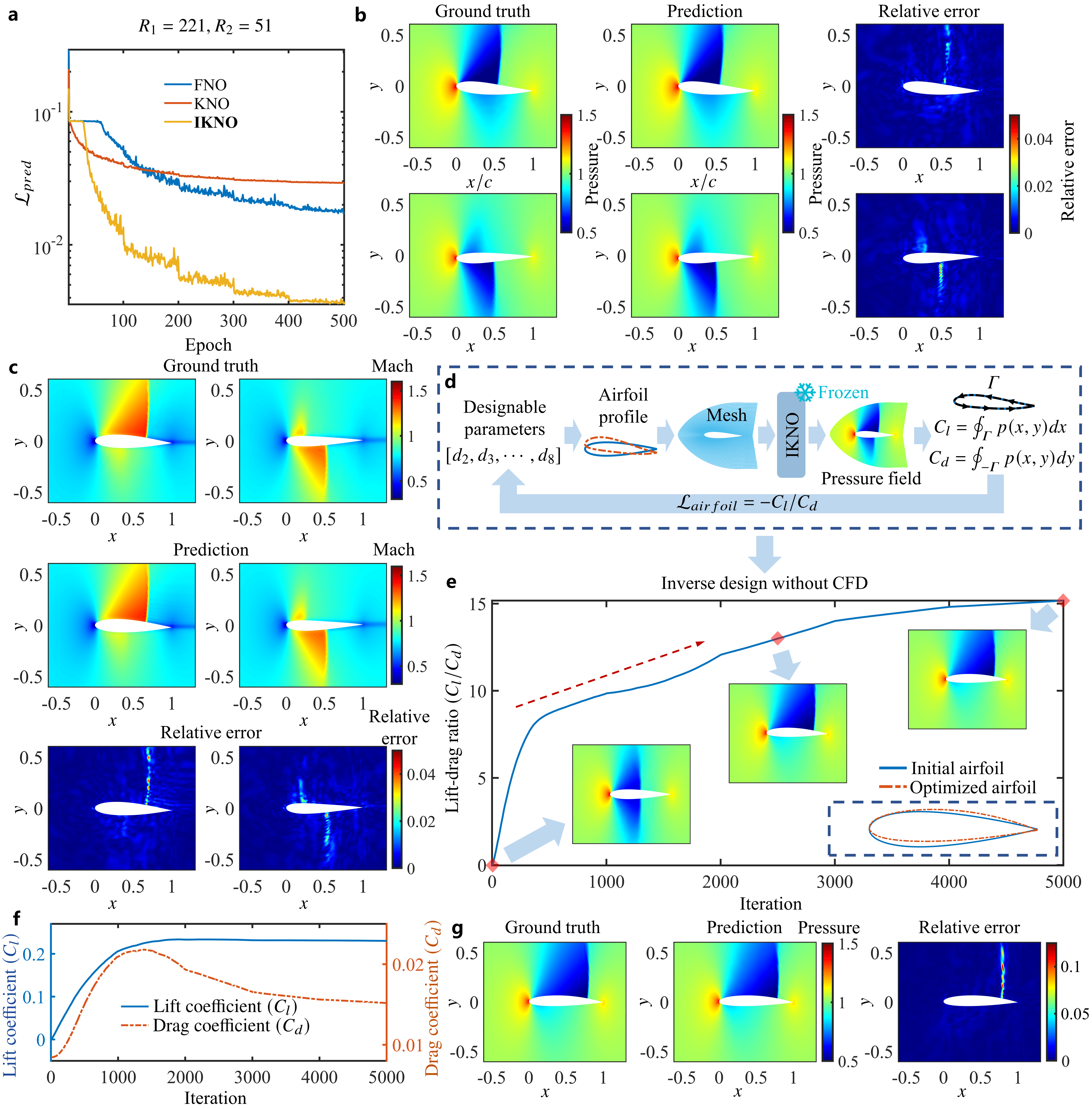}
      \caption{Results of the 2D compressible Euler equation for airfoil flow field, where the canonical coordinate map is utilized to transform the structured elliptic mesh to the uniform mesh defined on a Cartesian domain. \textbf{a} Comparison of the loss function. The proposed IKNO shows better convergence during training, achieving the lowest loss. \textbf{b} and \textbf{c} The ground truth, prediction, and relative error distribution of the pressure and velocity fields obtained by the IKNO with different airfoil shapes. \textbf{d} The trained IKNO is adopted for fast inverse design of airfoils without CFD. \textbf{e} Variation of lift-drug ratio with the number of iterations. \textbf{f} Variation of lift and drag coefficients with a number of iterations. \textbf{g} Comparison of predicted and CFD calculated pressure fields of the optimized airfoil.}
      \label{fig:airfoils}
\end{figure}

Under the settings above, Fig.\ref{fig:airfoils} presents the relevant results. Fig.\ref{fig:airfoils}a compares the loss function curves of different methods during training. We can observe that the proposed IKNO significantly outperforms KNO and FNO, and the loss function value at the end of training is almost an order of magnitude lower than the other two methods, showing its remarkable superiority in this challenging task. The prediction results of pressure and velocity fields for different airfoils in the test set are provided in Fig.\ref{fig:airfoils}b and c, respectively. It should be emphasized that the IKNO directly maps the airfoil profile information to pressure and velocity fields rather than through two separate models. It can be seen that the aerodynamic characteristics are complex and vary significantly between disparate airfoils. Far-field subsonic flow (0.8 Mach) changes to transonic behavior as it passes over the airfoil, leading to sharp changes in pressure and velocity fields near the trailing edge, indicating a shock wave. Away from the stagnation region, the pressure becomes negative (compared to the far-field pressure) due to the conservation of momentum. Given only the airfoil profile information, the proposed IKNO accurately predicts the pressure and velocity fields with an error of the order of $10^{-2}$, demonstrating its effectiveness. 

Moreover, since the IKNO is able to map the airfoil profile to the pressure field directly, we can further freeze the parameters of the trained IKNO and adopt it to guide the inverse design of the airfoil. As illustrated in Fig.\ref{fig:airfoils}c, the designable parameters determine the airfoil profile, which in turn allows for generating a structured elliptic mesh for solving the flow field, and subsequently, through the canonical coordinate transformation, the frozen IKNO can directly give the corresponding airfoil's pressure field. Subsequently, the lift and drag coefficients ($C_{l}$ and $C_{d}$) can be obtained from the path integral of the pressure field on the airfoil surface, which allows us to inverse design the airfoil profile with the objective of maximizing the lift-to-drag ratio (or equivalently, minimizing $-C_{l} / C_{d}$) through a gradient descent-based optimizer, e.g., Adam, like optimizing a neural network. The above process is very efficient because it does not involve Computational Fluid Dynamics (CFD), and corresponding results are plotted in Fig.\ref{fig:airfoils}e-g. The initial airfoil profile is symmetrical; hence, the lift coefficient is zero at an angle of attack of 0. As the optimization process proceeds, the airfoil becomes progressively asymmetrical, and the upper chamber of the upper edge gradually increases, leading to a larger negative pressure region, which helps the airfoil achieve a larger lift-to-drag ratio (shown in Fig.\ref{fig:airfoils}e) and lift coefficient (shown in Fig.\ref{fig:airfoils}f). At the end of the optimization process, compared with the initial airfoil (NACA0012), the lift-to-drag ratio is improved from 0 to 15.1721, $C_{L}$ is improved from 0 to 0.2303, and $C_{d}$ is kept at 0.0152, which achieves a satisfactory optimization result. Finally, we calculate the ground truth pressure field by CFD based on the optimized airfoil profile and compare it with that from the IKNO. As shown in Fig.\ref{fig:airfoils}g, they present excellent agreement, with most of the error concentrated near the shock wave and the vast majority of the other regions being very accurate. This result further fully demonstrates the IKNO's effectiveness and correctness in the inverse design task.

\subsection{A real-world example: global weather systems}

As our final example, we consider a more complex and real-world problem: global weather systems. The dataset is obtained from the European Centre for Medium-Range Weather Forecasts (ECMWF), which includes temperature data of air at 2 meters above the surface of the land, sea, or in-land waters from January 1, 2017, through April 21, 2022, for a total of 1937 days \cite{ERA5_url}. Measurements are taken at noon Universal Time Coordinated time each day, and the range is global, i.e., longitude $[-180^{\circ}, 180^{\circ}]$, latitude $[-90^{\circ}, 90^{\circ}]$. The resolution of original data is $0.25^{\circ} \times 0.25^{\circ}$, corresponding to $1440 \times 720$ mesh points. Here, we downsample them to the resolution of $360 \times 180$. Our goal is to learn a neural operator to make medium-range weather predictions, i.e., given the 2-meter temperature field at the initial $T_{d} = 7$ days, predict the 2-meter temperature field in the next week. The training and test sets are generated through a sliding window containing 270 and 5 samples, respectively. Moreover, to evaluate the resolution-invariance, the other test set with the resolution of $720 \times 360$ is also considered, which is $2\times$ the training set.

\begin{figure}[htbp]
      \centering
      \includegraphics[width = 0.97\textwidth]{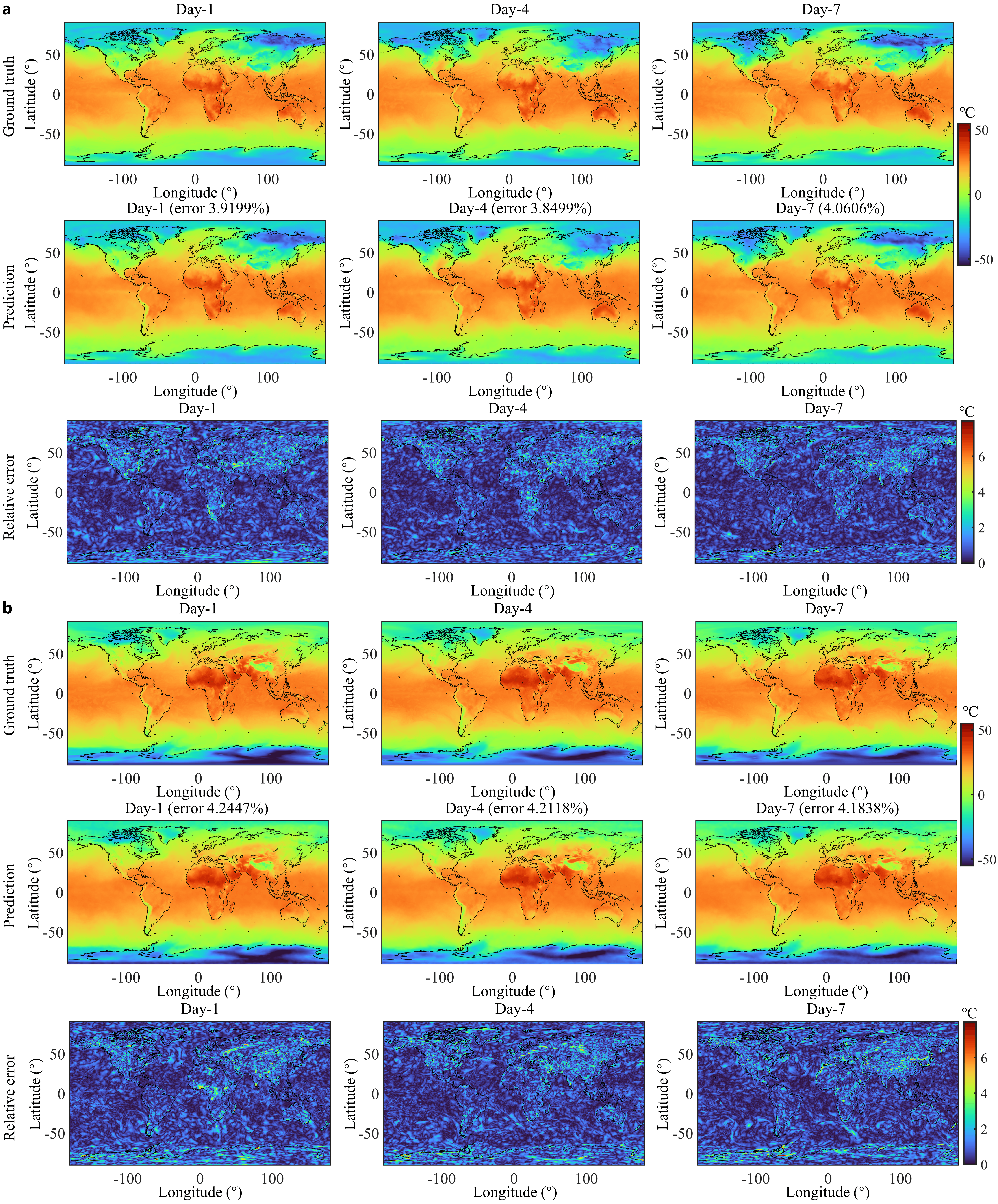}
      \caption{Results of the global weather systems. \textbf{a} The ground truth, prediction, and relative error distribution of the 2-meter temperature field obtained by the IKNO at the resolution of $360 \times 180$. \textbf{b} The ground truth, zero-shot super-resolution prediction, and relative error distribution of the 2-meter temperature field obtained by the IKNO at the resolution of $720 \times 360$, 2$\times$ the training set.}
      \label{fig:weather_temperature}
\end{figure}

Fig.\ref{fig:weather_temperature} summarizes the relevant results. As can be seen from Fig.\ref{fig:weather_temperature}a, the ground truth and predictions match well on a global scale of latitude and longitude, with $<5\%$ average relative error. Regarding the error's specific distribution, the vast majority of the global region is within $\pm 2^{\circ}$ of the error, proving the satisfactory match with the actual weather data. Moreover, the prediction results of higher resolution are provided in Fig.\ref{fig:weather_temperature}b, showing that the IKNO trained at $360 \times 180$ resolution can be directly used for higher resolution prediction with almost constant accuracy. These results further illustrate the effectiveness of the proposed IKNO in a real-world complex system, including data-driven modeling performance and resolution-invariance.

\section{Summaries and discussions}

Tab.\ref{tab:comparison} summarizes the comparison of data-driven modeling performance of different approaches in the discussed tasks. Two metrics on the test set are listed, i.e., MAE and relative $L_{2}$ error in percentage. Taken together, the proposed IKNO achieves the best performance in almost all test tasks, including predictions at zero-shot super-resolution conditions, only slightly worse than the best candidate in the super-resolution prediction problem for 2D Darcy flow. These results fully demonstrate the effectiveness and superiority of the proposed IKNO. Then, the necessary Fourier transforms in the IKNO are efficiently implemented through the FFT, but this treatment also implies that the input and output functions have to be defined in the Cartesian domain. However, by introducing some simple preprocessing strategies, such as interpolation and canonical coordinate map, the IKNO can be effectively applied to problems defined on non-Cartesian domains, as demonstrated in the 2D Darcy flow and airfoil flow field examples. Finally, it should be emphasized that the proposed IKNO is inspired by the Koopman operator theory, which describes the observable evolution on a time scale. Some typical problems in neural operators, such as predicting future states based on the initial time-delay snapshots, can be naturally incorporated into this framework, similar to the Koopman operator with time-delay coordinates. Many of the problems discussed in this paper, such as the 1D Burgers equation and 2D Navier-Stokes equivalently, fall into this category, and our interpretation of the IKNO's architecture also uses a similar perspective. However, some other problems, such as mapping boundary conditions to pressure fields for the 2D Darcy problem or mapping airfoil profiles to steady-state density, velocity, and pressure fields discussed in this paper, are different because the inputs and outputs are not intrinsically related through time advancement. But, interestingly, the results show that the IKNO still performs satisfactorily in these problems, suggesting that it can be utilized as a more generalized neural operator and is not limited to time-scale-involved problems. A possible explanation might be that the IKNO implicitly learns the underlying dynamics between the input and output functions, although this relationship may not explicitly exist in the original PDE formalism. This treatment is similar to the pseudo-time stepping format often adopted in CFD for solving steady-state problems, although these problems do not inherently vary with time. This perspective provides a new understanding of the IKNO's architecture and its potential applications in a broader range of problems.

\begin{table}[htbp]
      \caption{Comparison of data-driven modeling performance of different approaches in discussed tasks}
      \label{tab:comparison}
      \begin{tabular*}{1.0\textwidth}{@{}LLLLLLLLL@{}}
            \toprule
            \multirow{4}*{Example} & \multirow{4}{*}{\makecell[l]{Train \\ resolution}} & \multirow{4}{*}{\makecell[l]{Test \\ resolution}} & \multicolumn{2}{c}{FNO} & \multicolumn{2}{c}{KNO} & \multicolumn{2}{c}{\textbf{IKNO (ours)}} \\
            ~ & ~ & ~ & \multirow{3}{*}{MAE} & \multirow{3}{*}{\makecell[l]{Relavive \\ $L_{2}$ error \\ (\%)}} & \multirow{3}{*}{MAE} & \multirow{3}{*}{\makecell[l]{Relavive \\ $L_{2}$ error \\ (\%)}} & \multirow{3}{*}{MAE} & \multirow{3}{*}{\makecell[l]{Relavive \\ $L_{2}$ error \\ (\%)}} \\
            ~ & ~ & ~ & ~ & ~ & ~ & ~ & ~ & ~ \\
            ~ & ~ & ~ & ~ & ~ & ~ & ~ & ~ & ~ \\
            \midrule
            \multirow{6}*{1D Burgers equation} & \multirow{6}*{32} & 32 & 1.6316e-4 & 0.1296 & 2.1595e-4 & 0.1729 & \textbf{9.9118e-5} & \textbf{0.0824} \\
            ~ & ~ & 64 & 1.6234e-4 & 0.1280 & 5.4943e-4 & 0.5617 & \textbf{9.5072e-5} & \textbf{0.0763} \\
            ~ & ~ & 128 & 1.6227e-4 & 0.1279 & 5.4948e-4 & 0.5659 & \textbf{9.4659e-5} & \textbf{0.0759} \\
            ~ & ~ & 256 & 1.6228e-4 & 0.1279 & 5.4949e-4 & 0.5659 & \textbf{9.4529e-5} & \textbf{0.0757} \\
            ~ & ~ & 512 & 1.6225e-4 & 0.1279 & 5.4930e-4 & 0.5660 & \textbf{9.4435e-5} & \textbf{0.0757} \\
            ~ & ~ & 1024 & 1.6227e-4 & 0.1279 & 5.4933e-4 & 0.5660 & \textbf{9.4390e-5} & \textbf{0.0756} \\
            \midrule
            \multirow{2}*{2D shallow water equation} & \multirow{2}*{64$\times$64} & 64$\times$64 & 7.5963e-4 & 0.2031 & 2.1712e-3 & 0.6772 & \textbf{7.2205e-4} & \textbf{0.1864} \\
            ~ & ~ & 128$\times$128 & 8.0487e-4 & 0.2078 & 2.3850e-3 & 0.6782 & \textbf{7.9955e-4} & \textbf{0.1943} \\
            \midrule
            \multirow{2}*{\makecell[l]{2D incompressible \\ Navier-Stokes equation ($\nu = 10^{-3}$)}} & \multirow{2}*{64$\times$64} & \multirow{2}*{64$\times$64} & \multirow{2}*{7.2589e-3} & \multirow{2}*{1.0201} & \multirow{2}*{6.8984e-3} & \multirow{2}*{0.9590} & \multirow{2}*{\textbf{4.4748e-3}} & \multirow{2}*{\textbf{0.6213}} \\
            ~ & ~ & ~ & ~ & ~ & ~ & ~ & ~ & ~ \\
            \midrule
            \multirow{2}*{\makecell[l]{2D incompressible \\ Navier-Stokes equation ($\nu = 10^{-4}$)}} & \multirow{2}*{64$\times$64} & 64$\times$64 & 7.2844e-2 & 6.9542 & 1.4881e-1 & 13.6556 & \textbf{6.6610e-2} & \textbf{6.5059} \\
            ~ & ~ & 256$\times$256 & 7.2001e-2 & 6.9648 & 1.4594e-1 & 13.4826 & \textbf{6.9232e-2} & \textbf{6.9550} \\
            \midrule
            \multirow{2}*{2D Darcy flow} & \multirow{2}*{50$\times$50} & 50$\times$50 & 1.2711e-2 & 2.3162 & 8.1224e-3 & 1.4268 & \textbf{1.2949e-3} & \textbf{0.2264} \\
            ~ & ~ & 100$\times$100 & 1.9358e-2 & \textbf{3.5189} & \textbf{1.8882e-2} & 3.5896 & 1.9113e-2 & 3.5602 \\
            \midrule
            \multirow{2}*{\makecell[l]{2D compressible Euler \\ equation (interpolation)}} & \multirow{2}*{128$\times$128} & \multirow{2}*{128$\times$128} & \multirow{2}*{2.7511e-3} & \multirow{2}*{0.4858} & \multirow{2}*{2.3973e-2} & \multirow{2}*{3.8855} & \multirow{2}*{\textbf{2.2660e-3}} & \multirow{2}*{\textbf{0.3618}} \\
            ~ & ~ & ~ & ~ & ~ & ~ & ~ & ~ & ~ \\
            \midrule
            \multirow{3}*{\makecell[l]{2D compressible Euler \\ equation (canonical \\ coordinate map)}} & \multirow{3}*{221$\times$51} & \multirow{3}*{221$\times$51} & \multirow{3}*{5.7996e-3} & \multirow{3}*{1.8353} & \multirow{3}*{1.2832e-2} & \multirow{3}*{3.0310} & \multirow{3}*{\textbf{1.9298e-3}} & \multirow{3}*{\textbf{0.5329}} \\
            ~ & ~ & ~ & ~ & ~ & ~ & ~ & ~ & ~ \\
            ~ & ~ & ~ & ~ & ~ & ~ & ~ & ~ & ~ \\
            \midrule
            \multirow{2}*{Global weather systems} & \multirow{2}*{360$\times$180} & 360$\times$180 & 1.5112e0 & 9.4467 & 1.6052e0 & 9.8718 & \textbf{7.3004e-1} & \textbf{4.3802} \\
            ~ & ~ & 720$\times$360 & 1.5165e0 & 9.5215 & 1.6100e0 & 9.9373 & \textbf{7.8598e-1} & \textbf{4.7827} \\
            \bottomrule
      \end{tabular*}
\end{table}

\section{Conclusions and future works}

This paper develops a novel data-driven method for PDEs, denoted as IKNO. The main conclusions are as follows:

(1) IKNO is inspired by the Koopman operator theory and neural operator, and each component is designed and introduced with a clear motivation and mathematical correspondence.

(2) The reconstruction relation is explicitly guaranteed in IKNO by simultaneously parameterizing the observable function and its inverse through the INN, removing the dependence on the reconstruction loss. This novel treatment is an essential improvement over the original KNO.

(3) The structured linear matrix adopted to approximate the Koopmna operator is parameterized in the frequency space rather than directly in the observable space to learn the evolution of the observables' low-frequency modes, sustaining IKNO is resolution-invariant like other neural operators.

(4) Introducing FFT and IFFT enables efficient Fourier and inverse Fourier transforms but restricts the input function's form, which must be defined on the Cartesian domain. Fortunately, by introducing some simple preprocessing strategies, such as interpolation and canonical coordinate map, the IKNO can be effectively applied to problems defined on non-Cartesian domains.

(5) The proposed IKNO is validated and compared with FNO and KNO in a series of tasks, including the 1D Burgers equation, the 2D shallow water equation, the 2D incompressible Navier-Stokes equation, the 2D Darcy flow, the 2D compressible Euler equation for the airfoil flow field, and global weather systems. The results show that the IKNO performs best in almost all test tasks, including predictions at zero-shot super-resolution conditions.

Future works will focus on further improvements and applications of IKNO. For example, enhancement of current IKNO based on the factorization \cite{tranFactorizedFourierNeural2023} or U-Net-like architecture \cite{liuMSIUFFNOMultiscaleImplicit2025}. Alternatively, adopting the direct spectral evaluations \cite{lingschRegularGridsFourierBased2024} to make the IKNO suitable for more complex geometries. It also makes sense to combine IKNO as a branch network with DeepONet, thus allowing input and output functions to be defined in different domains \cite{huangResolutionInvariantDeep2025a}. Finally, establishing digital twin models of complex systems based on IKNO will be interesting and meaningful. 

\section*{Acknowledgements}
It is very grateful for the National Key R \& D Program of China (Grant No. 2023YFE0125900), National Natural Science Foundation of China (Grant Nos. 12422213, U244120491, and 12372008), the key research and development project of Heilongjiang Province (Grant No. 2023ZX01A03), and the Stabilization Support Project for Basic Military Research Institutes (Grant No. 03020051).













\printcredits

\bibliographystyle{model1-num-names}

\bibliography{cas-refs}

\begin{thebibliography}{115}
\expandafter\ifx\csname natexlab\endcsname\relax\def\natexlab#1{#1}\fi
\providecommand{\url}[1]{\texttt{#1}}
\providecommand{\href}[2]{#2}
\providecommand{\path}[1]{#1}
\providecommand{\DOIprefix}{doi:}
\providecommand{\ArXivprefix}{arXiv:}
\providecommand{\URLprefix}{URL: }
\providecommand{\Pubmedprefix}{pmid:}
\providecommand{\doi}[1]{\href{http://dx.doi.org/#1}{\path{#1}}}
\providecommand{\Pubmed}[1]{\href{pmid:#1}{\path{#1}}}
\providecommand{\bibinfo}[2]{#2}
\ifx\xfnm\relax \def\xfnm[#1]{\unskip,\space#1}\fi
\bibitem[{Ren et~al.(2025)Ren, Hou, Yuan, Duraihem, Awwad, and
  Saeed}]{REN2025119023}
\bibinfo{author}{S.~Ren}, \bibinfo{author}{L.~Hou}, \bibinfo{author}{T.~Yuan},
  \bibinfo{author}{F.~Z. Duraihem}, \bibinfo{author}{E.~M. Awwad},
  \bibinfo{author}{N.~Saeed},
\newblock \bibinfo{title}{A novel formulation for efficient flutter analysis of
  rotating composite blades based on referenced nodal coordinate formulation},
\newblock \bibinfo{journal}{Composite Structures} \bibinfo{volume}{359}
  (\bibinfo{year}{2025}) \bibinfo{pages}{119023}.
\bibitem[{Xu et~al.(2025)Xu, Hou, Hou, Li, Ren, Duraihem, {Mahrous Awwad}, and
  Saeed}]{XU2025109961}
\bibinfo{author}{Q.~Xu}, \bibinfo{author}{L.~Hou}, \bibinfo{author}{L.~Hou},
  \bibinfo{author}{Z.~Li}, \bibinfo{author}{S.~Ren}, \bibinfo{author}{F.~Z.
  Duraihem}, \bibinfo{author}{E.~{Mahrous Awwad}}, \bibinfo{author}{N.~A.
  Saeed},
\newblock \bibinfo{title}{A novel {ROM}-based {FSI} model of composite blisk
  with blades-disk coupling for flutter analysis},
\newblock \bibinfo{journal}{Aerospace Science and Technology}
  \bibinfo{volume}{159} (\bibinfo{year}{2025}) \bibinfo{pages}{109961}.
\bibitem[{Meng et~al.(2025)Meng, Hou, Wang, Lin, Li, Zhong, Chen, Saeed,
  Mohamed, and Awwad}]{MENG2025118674}
\bibinfo{author}{Q.~Meng}, \bibinfo{author}{L.~Hou}, \bibinfo{author}{A.~Wang},
  \bibinfo{author}{R.~Lin}, \bibinfo{author}{Z.~Li},
  \bibinfo{author}{S.~Zhong}, \bibinfo{author}{Y.~Chen}, \bibinfo{author}{N.~A.
  Saeed}, \bibinfo{author}{A.~Mohamed}, \bibinfo{author}{E.~Awwad},
\newblock \bibinfo{title}{Subharmonic response suppression of a quasi-zero
  stiffness system},
\newblock \bibinfo{journal}{Journal of Sound and Vibration}
  \bibinfo{volume}{594} (\bibinfo{year}{2025}) \bibinfo{pages}{118674}.
\bibitem[{Wang et~al.(2023)Wang, Li, Gong, and Xuan}]{WANG2023109242}
\bibinfo{author}{H.~Wang}, \bibinfo{author}{B.~Li}, \bibinfo{author}{J.~Gong},
  \bibinfo{author}{F.-Z. Xuan},
\newblock \bibinfo{title}{Machine learning-based fatigue life prediction of
  metal materials: Perspectives of physics-informed and data-driven hybrid
  methods},
\newblock \bibinfo{journal}{Engineering Fracture Mechanics}
  \bibinfo{volume}{284} (\bibinfo{year}{2023}) \bibinfo{pages}{109242}.
\bibitem[{North et~al.(2023)North, Wikle, and
  Schliep}]{northReviewDataDrivenDiscovery2023}
\bibinfo{author}{J.~S. North}, \bibinfo{author}{C.~K. Wikle},
  \bibinfo{author}{E.~M. Schliep},
\newblock \bibinfo{title}{A {{Review}} of {{Data}}-{{Driven Discovery}} for
  {{Dynamic Systems}}},
\newblock \bibinfo{journal}{International Statistical Review}
  \bibinfo{volume}{91} (\bibinfo{year}{2023}) \bibinfo{pages}{464--492}.
\bibitem[{Ye et~al.(2025)Ye, Li, Liu, Xiao, and
  Wan}]{yeOverviewDataDrivenModels2025}
\bibinfo{author}{M.~Ye}, \bibinfo{author}{M.~Li}, \bibinfo{author}{M.~Liu},
  \bibinfo{author}{C.~Xiao}, \bibinfo{author}{D.~Wan},
\newblock \bibinfo{title}{Overview of {{Data-Driven Models}} for {{Wind Turbine
  Wake Flows}}},
\newblock \bibinfo{journal}{Journal of Marine Science and Application}
  \bibinfo{volume}{24} (\bibinfo{year}{2025}) \bibinfo{pages}{1--20}.
\bibitem[{Yang et~al.(2023)Yang, Liu, Meng, and Osher}]{Yang_2023}
\bibinfo{author}{L.~Yang}, \bibinfo{author}{S.~Liu}, \bibinfo{author}{T.~Meng},
  \bibinfo{author}{S.~J. Osher},
\newblock \bibinfo{title}{In-context operator learning with data prompts for
  differential equation problems},
\newblock \bibinfo{journal}{Proceedings of the National Academy of Sciences}
  \bibinfo{volume}{120} (\bibinfo{year}{2023}).
\bibitem[{Na et~al.(2023)Na, Ren, Liu, and Han}]{9741840}
\bibinfo{author}{X.~Na}, \bibinfo{author}{W.~Ren}, \bibinfo{author}{M.~Liu},
  \bibinfo{author}{M.~Han},
\newblock \bibinfo{title}{Hierarchical echo state network with sparse learning:
  A method for multidimensional chaotic time series prediction},
\newblock \bibinfo{journal}{IEEE Transactions on Neural Networks and Learning
  Systems} \bibinfo{volume}{34} (\bibinfo{year}{2023})
  \bibinfo{pages}{9302--9313}.
\bibitem[{Hu et~al.(2021)Hu, Wang, and Tao}]{HU2021729}
\bibinfo{author}{H.~Hu}, \bibinfo{author}{L.~Wang}, \bibinfo{author}{R.~Tao},
\newblock \bibinfo{title}{Wind speed forecasting based on variational mode
  decomposition and improved echo state network},
\newblock \bibinfo{journal}{Renewable Energy} \bibinfo{volume}{164}
  (\bibinfo{year}{2021}) \bibinfo{pages}{729--751}.
\bibitem[{Brunton et~al.(2016)Brunton, Proctor, and Kutz}]{SINDy}
\bibinfo{author}{S.~L. Brunton}, \bibinfo{author}{J.~L. Proctor},
  \bibinfo{author}{J.~N. Kutz},
\newblock \bibinfo{title}{Discovering governing equations from data by sparse
  identification of nonlinear dynamical systems},
\newblock \bibinfo{journal}{Proceedings of the National Academy of Sciences of
  the United States of America} \bibinfo{volume}{113} (\bibinfo{year}{2016})
  \bibinfo{pages}{3932--3937}.
\bibitem[{Kaise et~al.(2018)Kaise, Kutz, and Brunton}]{SINDyc}
\bibinfo{author}{E.~Kaise}, \bibinfo{author}{J.~N. Kutz},
  \bibinfo{author}{S.~L. Brunton},
\newblock \bibinfo{title}{Sparse identification of nonlinear dynamics for model
  predictive control in the low-data limit},
\newblock \bibinfo{journal}{Proceedings of the Royal Society A}
  \bibinfo{volume}{474} (\bibinfo{year}{2018}) \bibinfo{pages}{20180335}.
\bibitem[{Naozuka et~al.(2022)Naozuka, Rocha, Silva, and
  Almeida}]{naozukaSINDySAFrameworkEnhancing2022}
\bibinfo{author}{G.~T. Naozuka}, \bibinfo{author}{H.~L. Rocha},
  \bibinfo{author}{R.~S. Silva}, \bibinfo{author}{R.~C. Almeida},
\newblock \bibinfo{title}{Sindy-sa framework: Enhancing nonlinear system
  identification with sensitivity analysis},
\newblock \bibinfo{journal}{Nonlinear Dynamics} \bibinfo{volume}{110}
  (\bibinfo{year}{2022}) \bibinfo{pages}{2589--2609}.
\bibitem[{Champion et~al.(2019)Champion, Lusch, Kutz, and
  Brunton}]{doi:10.1073/pnas.1906995116}
\bibinfo{author}{K.~Champion}, \bibinfo{author}{B.~Lusch},
  \bibinfo{author}{J.~N. Kutz}, \bibinfo{author}{S.~L. Brunton},
\newblock \bibinfo{title}{Data-driven discovery of coordinates and governing
  equations},
\newblock \bibinfo{journal}{Proceedings of the National Academy of Sciences}
  \bibinfo{volume}{116} (\bibinfo{year}{2019}) \bibinfo{pages}{22445--22451}.
\bibitem[{Cenedese et~al.(2022)Cenedese, Ax{\aa}s, B{\"a}uerlein, Avila, and
  Haller}]{cenedeseDatadrivenModelingPrediction2022}
\bibinfo{author}{M.~Cenedese}, \bibinfo{author}{J.~Ax{\aa}s},
  \bibinfo{author}{B.~B{\"a}uerlein}, \bibinfo{author}{K.~Avila},
  \bibinfo{author}{G.~Haller},
\newblock \bibinfo{title}{Data-driven modeling and prediction of
  non-linearizable dynamics via spectral submanifolds},
\newblock \bibinfo{journal}{Nature Communications} \bibinfo{volume}{13}
  (\bibinfo{year}{2022}) \bibinfo{pages}{872}.
\bibitem[{Alora et~al.(2023)Alora, Cenedese, Schmerling, Haller, and
  Pavone}]{10160418}
\bibinfo{author}{J.~I. Alora}, \bibinfo{author}{M.~Cenedese},
  \bibinfo{author}{E.~Schmerling}, \bibinfo{author}{G.~Haller},
  \bibinfo{author}{M.~Pavone},
\newblock \bibinfo{title}{Data-driven spectral submanifold reduction for
  nonlinear optimal control of high-dimensional robots},
\newblock in: \bibinfo{booktitle}{2023 IEEE International Conference on
  Robotics and Automation (ICRA)}, \bibinfo{year}{2023}, pp.
  \bibinfo{pages}{2627--2633}.
\bibitem[{Ax{\aa}s et~al.(2023)Ax{\aa}s, Cenedese, and
  Haller}]{axasFastDatadrivenModel2023}
\bibinfo{author}{J.~Ax{\aa}s}, \bibinfo{author}{M.~Cenedese},
  \bibinfo{author}{G.~Haller},
\newblock \bibinfo{title}{Fast data-driven model reduction for nonlinear
  dynamical systems},
\newblock \bibinfo{journal}{Nonlinear Dynamics} \bibinfo{volume}{111}
  (\bibinfo{year}{2023}) \bibinfo{pages}{7941--7957}.
\bibitem[{Parmar et~al.(2022)Parmar, Refai, and Runolfsson}]{9037083}
\bibinfo{author}{N.~Parmar}, \bibinfo{author}{H.~H. Refai},
  \bibinfo{author}{T.~Runolfsson},
\newblock \bibinfo{title}{A survey on the methods and results of data-driven
  koopman analysis in the visualization of dynamical systems},
\newblock \bibinfo{journal}{IEEE Transactions on Big Data} \bibinfo{volume}{8}
  (\bibinfo{year}{2022}) \bibinfo{pages}{723--738}.
\bibitem[{Otto and Rowley(2019)}]{doi:10.1137/18M1177846}
\bibinfo{author}{S.~E. Otto}, \bibinfo{author}{C.~W. Rowley},
\newblock \bibinfo{title}{Linearly recurrent autoencoder networks for learning
  dynamics},
\newblock \bibinfo{journal}{SIAM Journal on Applied Dynamical Systems}
  \bibinfo{volume}{18} (\bibinfo{year}{2019}) \bibinfo{pages}{558--593}.
\bibitem[{Lusch et~al.(2018)Lusch, Kutz, and Brunton}]{DeepKoopman_Lusch}
\bibinfo{author}{B.~Lusch}, \bibinfo{author}{J.~N. Kutz},
  \bibinfo{author}{S.~L. Brunton},
\newblock \bibinfo{title}{Deep learning for universal linear embeddings of
  nonlinear dynamics},
\newblock \bibinfo{journal}{Nature Communications} \bibinfo{volume}{9}
  (\bibinfo{year}{2018}) \bibinfo{pages}{4950}.
\bibitem[{Gao et~al.(2025)Gao, Williams, and
  Kutz}]{gao2025sparseidentificationnonlineardynamics}
\bibinfo{author}{M.~L. Gao}, \bibinfo{author}{J.~P. Williams},
  \bibinfo{author}{J.~N. Kutz}, \bibinfo{title}{Sparse identification of
  nonlinear dynamics and koopman operators with shallow recurrent decoder
  networks}, \bibinfo{year}{2025}.
  \DOIprefix\doi{https://arxiv.org/abs/2501.13329}.
  \href{http://arxiv.org/abs/arXiv: 2501.13329}{\tt arXiv:arXiv: 2501.13329}.
\bibitem[{Wilson(2023)}]{wilsonKoopmanOperatorInspired2023}
\bibinfo{author}{D.~Wilson},
\newblock \bibinfo{title}{Koopman {{Operator Inspired Nonlinear System
  Identification}}},
\newblock \bibinfo{journal}{SIAM Journal on Applied Dynamical Systems}
  \bibinfo{volume}{22} (\bibinfo{year}{2023}) \bibinfo{pages}{1445--1471}.
\bibitem[{Koopman(1931)}]{koopmanHamiltonianSystemsTransformation1931}
\bibinfo{author}{B.~O. Koopman},
\newblock \bibinfo{title}{Hamiltonian systems and transformation in hilbert
  space},
\newblock \bibinfo{journal}{Proceedings of the National Academy of Sciences}
  \bibinfo{volume}{17} (\bibinfo{year}{1931}) \bibinfo{pages}{315--318}.
\bibitem[{H.~Tu et~al.(2014)H.~Tu, W.~Rowley, M.~Luchtenburg, L.~Brunton, and
  Nathan~Kutz}]{H_Tu_2014}
\bibinfo{author}{J.~H.~Tu}, \bibinfo{author}{C.~W.~Rowley},
  \bibinfo{author}{D.~M.~Luchtenburg}, \bibinfo{author}{S.~L.~Brunton},
  \bibinfo{author}{J.~Nathan~Kutz},
\newblock \bibinfo{title}{On dynamic mode decomposition: Theory and
  applications},
\newblock \bibinfo{journal}{Journal of Computational Dynamics}
  \bibinfo{volume}{1} (\bibinfo{year}{2014}) \bibinfo{pages}{391--421}.
\bibitem[{Schmid(2010)}]{doi.org/10.1017/S0022112009992059}
\bibinfo{author}{P.~J. Schmid},
\newblock \bibinfo{title}{Dynamic mode decomposition of numerical and
  experimental data},
\newblock \bibinfo{journal}{Journal of Fluid Mechanics} \bibinfo{volume}{656}
  (\bibinfo{year}{2010}) \bibinfo{pages}{5--28}.
\bibitem[{Williams et~al.(2015)Williams, Kevrekidis, and Rowley}]{EDMD}
\bibinfo{author}{M.~O. Williams}, \bibinfo{author}{I.~G. Kevrekidis},
  \bibinfo{author}{C.~W. Rowley},
\newblock \bibinfo{title}{A data-driven approximation of the koopman operator:
  Extending dynamic mode decomposition},
\newblock \bibinfo{journal}{Journal of Nonlinear Science} \bibinfo{volume}{25}
  (\bibinfo{year}{2015}) \bibinfo{pages}{1307--1346}.
\bibitem[{O.~Williams et~al.(2015)O.~Williams, W.~Rowley, and
  G.~Kevrekidis}]{o.williamsKernelbasedMethodDatadriven2015}
\bibinfo{author}{M.~O.~Williams}, \bibinfo{author}{C.~W.~Rowley},
  \bibinfo{author}{I.~G.~Kevrekidis},
\newblock \bibinfo{title}{A kernel-based method for data-driven koopman
  spectral analysis},
\newblock \bibinfo{journal}{Journal of Computational Dynamics}
  \bibinfo{volume}{2} (\bibinfo{year}{2015}) \bibinfo{pages}{247--265}.
\bibitem[{Arbabi and Mezi{\'c}(2017)}]{Arbabi_2017}
\bibinfo{author}{H.~Arbabi}, \bibinfo{author}{I.~Mezi{\'c}},
\newblock \bibinfo{title}{Ergodic theory, dynamic mode decomposition, and
  computation of spectral properties of the koopman operator},
\newblock \bibinfo{journal}{SIAM Journal on Applied Dynamical Systems}
  \bibinfo{volume}{16} (\bibinfo{year}{2017}) \bibinfo{pages}{2096--2126}.
\bibitem[{Le~Clainche and Vega(2017)}]{leclaincheHigherOrderDynamic2017}
\bibinfo{author}{S.~Le~Clainche}, \bibinfo{author}{J.~M. Vega},
\newblock \bibinfo{title}{Higher order dynamic mode decomposition},
\newblock \bibinfo{journal}{SIAM Journal on Applied Dynamical Systems}
  \bibinfo{volume}{16} (\bibinfo{year}{2017}) \bibinfo{pages}{882--925}.
\bibitem[{Kutz et~al.(2016)Kutz, Fu, and Brunton}]{doi:10.1137/15M1023543}
\bibinfo{author}{J.~N. Kutz}, \bibinfo{author}{X.~Fu}, \bibinfo{author}{S.~L.
  Brunton},
\newblock \bibinfo{title}{Multiresolution dynamic mode decomposition},
\newblock \bibinfo{journal}{SIAM Journal on Applied Dynamical Systems}
  \bibinfo{volume}{15} (\bibinfo{year}{2016}) \bibinfo{pages}{713--735}.
\bibitem[{Jovanovi{\'c} et~al.(2014)Jovanovi{\'c}, Schmid, and
  Nichols}]{jovanovicSparsitypromotingDynamicMode2014}
\bibinfo{author}{M.~R. Jovanovi{\'c}}, \bibinfo{author}{P.~J. Schmid},
  \bibinfo{author}{J.~W. Nichols},
\newblock \bibinfo{title}{Sparsity-promoting dynamic mode decomposition},
\newblock \bibinfo{journal}{Physics of Fluids} \bibinfo{volume}{26}
  (\bibinfo{year}{2014}) \bibinfo{pages}{024103}.
\bibitem[{Hemati et~al.(2017)Hemati, Rowley, Deem, and
  Cattafesta}]{hematiDebiasingDynamicMode2017}
\bibinfo{author}{M.~S. Hemati}, \bibinfo{author}{C.~W. Rowley},
  \bibinfo{author}{E.~A. Deem}, \bibinfo{author}{L.~N. Cattafesta},
\newblock \bibinfo{title}{De-biasing the dynamic mode decomposition for applied
  {{Koopman}} spectral analysis of noisy datasets},
\newblock \bibinfo{journal}{Theoretical and Computational Fluid Dynamics}
  \bibinfo{volume}{31} (\bibinfo{year}{2017}) \bibinfo{pages}{349--368}.
\bibitem[{Dawson et~al.(2016)Dawson, Hemati, Williams, and
  Rowley}]{dawsonCharacterizingCorrectingEffect2016}
\bibinfo{author}{S.~T.~M. Dawson}, \bibinfo{author}{M.~S. Hemati},
  \bibinfo{author}{M.~O. Williams}, \bibinfo{author}{C.~W. Rowley},
\newblock \bibinfo{title}{Characterizing and correcting for the effect of
  sensor noise in the dynamic mode decomposition},
\newblock \bibinfo{journal}{Experiments in Fluids} \bibinfo{volume}{57}
  (\bibinfo{year}{2016}) \bibinfo{pages}{42}.
\bibitem[{Baddoo et~al.(2023)Baddoo, Herrmann, McKeon, Nathan~Kutz, and
  Brunton}]{baddooPhysicsinformedDynamicMode2023a}
\bibinfo{author}{P.~J. Baddoo}, \bibinfo{author}{B.~Herrmann},
  \bibinfo{author}{B.~J. McKeon}, \bibinfo{author}{J.~Nathan~Kutz},
  \bibinfo{author}{S.~L. Brunton},
\newblock \bibinfo{title}{Physics-informed dynamic mode decomposition},
\newblock \bibinfo{journal}{Proceedings of the Royal Society A: Mathematical,
  Physical and Engineering Sciences} \bibinfo{volume}{479}
  (\bibinfo{year}{2023}) \bibinfo{pages}{20220576}.
\bibitem[{Haller and
  Kasz{\'a}s(2024)}]{hallerDatadrivenLinearizationDynamical2024}
\bibinfo{author}{G.~Haller}, \bibinfo{author}{B.~Kasz{\'a}s},
\newblock \bibinfo{title}{Data-driven linearization of dynamical systems},
\newblock \bibinfo{journal}{Nonlinear Dynamics} \bibinfo{volume}{112}
  (\bibinfo{year}{2024}) \bibinfo{pages}{18639--18663}.
\bibitem[{Haseli and
  Cort{\'e}s(2022)}]{haseliLearningKoopmanEigenfunctions2022}
\bibinfo{author}{M.~Haseli}, \bibinfo{author}{J.~Cort{\'e}s},
\newblock \bibinfo{title}{Learning koopman eigenfunctions and invariant
  subspaces from data: Symmetric subspace decomposition},
\newblock \bibinfo{journal}{IEEE Transactions on Automatic Control}
  \bibinfo{volume}{67} (\bibinfo{year}{2022}) \bibinfo{pages}{3442--3457}.
\bibitem[{Li and Utyuzhnikov(2024)}]{liPredictionWindEnergy2024}
\bibinfo{author}{K.~Li}, \bibinfo{author}{S.~Utyuzhnikov},
\newblock \bibinfo{title}{Prediction of wind energy with the use of
  tensor-train based higher order dynamic mode decomposition},
\newblock \bibinfo{journal}{Journal of Forecasting}  (\bibinfo{year}{2024})
  \bibinfo{pages}{for.3126}.
\bibitem[{Endo et~al.(2024)Endo, Ikeda, Harada, Yamagata, Matsubara, Matsuo,
  Kawahara, and Yamashita}]{endoManifoldAlterationMajor2024}
\bibinfo{author}{H.~Endo}, \bibinfo{author}{S.~Ikeda},
  \bibinfo{author}{K.~Harada}, \bibinfo{author}{H.~Yamagata},
  \bibinfo{author}{T.~Matsubara}, \bibinfo{author}{K.~Matsuo},
  \bibinfo{author}{Y.~Kawahara}, \bibinfo{author}{O.~Yamashita},
\newblock \bibinfo{title}{Manifold alteration between major depressive disorder
  and healthy control subjects using dynamic mode decomposition in
  resting-state {{fMRI}} data},
\newblock \bibinfo{journal}{Frontiers in Psychiatry} \bibinfo{volume}{15}
  (\bibinfo{year}{2024}) \bibinfo{pages}{1288808}.
\bibitem[{Lortie and Forbes(2024)}]{lortieAsymptoticallyStableDataDriven2024}
\bibinfo{author}{L.~Lortie}, \bibinfo{author}{J.~R. Forbes},
  \bibinfo{title}{Asymptotically {{Stable Data-Driven Koopman Operator
  Approximation}} with {{Inputs}} using {{Total Extended DMD}}},
  \bibinfo{year}{2024}. \DOIprefix\doi{https://arxiv.org/abs/2408.16846}.
  \href{http://arxiv.org/abs/arXiv: 2408.16846}{\tt arXiv:arXiv: 2408.16846}.
\bibitem[{Leventides et~al.(2022)Leventides, Melas, and
  Poulios}]{leventidesExtendedDynamicMode2022}
\bibinfo{author}{J.~Leventides}, \bibinfo{author}{E.~Melas},
  \bibinfo{author}{C.~Poulios},
\newblock \bibinfo{title}{Extended dynamic mode decomposition for cyclic
  macroeconomic data},
\newblock \bibinfo{journal}{Data Science in Finance and Economics}
  \bibinfo{volume}{2} (\bibinfo{year}{2022}) \bibinfo{pages}{117--146}.
\bibitem[{Mendez et~al.(2021)Mendez, Le~Clainche, {Moreno-Ramos}, and
  Vega}]{mendezNewAutomaticVery2021}
\bibinfo{author}{C.~Mendez}, \bibinfo{author}{S.~Le~Clainche},
  \bibinfo{author}{R.~{Moreno-Ramos}}, \bibinfo{author}{J.~M. Vega},
\newblock \bibinfo{title}{A new automatic, very efficient method for the
  analysis of flight flutter testing data},
\newblock \bibinfo{journal}{Aerospace Science and Technology}
  \bibinfo{volume}{114} (\bibinfo{year}{2021}) \bibinfo{pages}{106749}.
\bibitem[{Ma et~al.(2022)Ma, Zhang, and Wang}]{maAdaptiveDynamicMode2022}
\bibinfo{author}{P.~Ma}, \bibinfo{author}{H.~Zhang}, \bibinfo{author}{C.~Wang},
\newblock \bibinfo{title}{Adaptive dynamic mode decomposition and its
  application in rolling bearing compound fault diagnosis},
\newblock \bibinfo{journal}{Structural Health Monitoring}
  (\bibinfo{year}{2022}) \bibinfo{pages}{147592172210957}.
\bibitem[{Iacob et~al.(2024)Iacob, T{\'o}th, and
  Schoukens}]{iacobKoopmanFormNonlinear2024}
\bibinfo{author}{L.~C. Iacob}, \bibinfo{author}{R.~T{\'o}th},
  \bibinfo{author}{M.~Schoukens},
\newblock \bibinfo{title}{Koopman form of nonlinear systems with inputs},
\newblock \bibinfo{journal}{Automatica} \bibinfo{volume}{162}
  (\bibinfo{year}{2024}) \bibinfo{pages}{111525}.
\bibitem[{Ren et~al.(2024)Ren, Fan, Bai, Ma, and
  Zhao}]{renPredictionSpatiotemporalDynamic2024}
\bibinfo{author}{H.~Ren}, \bibinfo{author}{M.~Fan}, \bibinfo{author}{Y.~Bai},
  \bibinfo{author}{X.~Ma}, \bibinfo{author}{J.~Zhao},
\newblock \bibinfo{title}{Prediction of spatiotemporal dynamic systems by
  data-driven reconstruction},
\newblock \bibinfo{journal}{Chaos, Solitons \& Fractals} \bibinfo{volume}{185}
  (\bibinfo{year}{2024}) \bibinfo{pages}{115137}.
\bibitem[{Folkestad et~al.(2020)Folkestad, Pastor, and Burdick}]{9197510}
\bibinfo{author}{C.~Folkestad}, \bibinfo{author}{D.~Pastor},
  \bibinfo{author}{J.~W. Burdick},
\newblock \bibinfo{title}{Episodic {Koopman} learning of nonlinear robot
  dynamics with application to fast multirotor landing},
\newblock in: \bibinfo{booktitle}{2020 IEEE International Conference on
  Robotics and Automation (ICRA)}, \bibinfo{year}{2020}, pp.
  \bibinfo{pages}{9216--9222}. \DOIprefix\doi{10.1109/ICRA40945.2020.9197510}.
\bibitem[{Mamakoukas et~al.(2023)Mamakoukas, Abraham, and Murphey}]{10091950}
\bibinfo{author}{G.~Mamakoukas}, \bibinfo{author}{I.~Abraham},
  \bibinfo{author}{T.~D. Murphey},
\newblock \bibinfo{title}{Learning stable models for prediction and control},
\newblock \bibinfo{journal}{IEEE Transactions on Robotics} \bibinfo{volume}{39}
  (\bibinfo{year}{2023}) \bibinfo{pages}{2255--2275}.
\bibitem[{Mamakoukas et~al.(2021)Mamakoukas, Castaño, Tan, and
  Murphey}]{9442852}
\bibinfo{author}{G.~Mamakoukas}, \bibinfo{author}{M.~L. Castaño},
  \bibinfo{author}{X.~Tan}, \bibinfo{author}{T.~D. Murphey},
\newblock \bibinfo{title}{Derivative-based koopman operators for real-time
  control of robotic systems},
\newblock \bibinfo{journal}{IEEE Transactions on Robotics} \bibinfo{volume}{37}
  (\bibinfo{year}{2021}) \bibinfo{pages}{2173--2192}.
\bibitem[{Klus et~al.(2020)Klus, N{\"u}ske, Peitz, Niemann, Clementi, and
  Sch{\"u}tte}]{klusDatadrivenApproximationKoopman2020}
\bibinfo{author}{S.~Klus}, \bibinfo{author}{F.~N{\"u}ske},
  \bibinfo{author}{S.~Peitz}, \bibinfo{author}{J.-H. Niemann},
  \bibinfo{author}{C.~Clementi}, \bibinfo{author}{C.~Sch{\"u}tte},
\newblock \bibinfo{title}{Data-driven approximation of the {Koopman} generator:
  Model reduction, system identification, and control},
\newblock \bibinfo{journal}{Physica D: Nonlinear Phenomena}
  \bibinfo{volume}{406} (\bibinfo{year}{2020}) \bibinfo{pages}{132416}.
\bibitem[{Han et~al.(2025)Han, Peng, Chen, and
  Liu}]{hanDatadrivenKoopmanModeling2025}
\bibinfo{author}{L.~Han}, \bibinfo{author}{K.~Peng}, \bibinfo{author}{W.~Chen},
  \bibinfo{author}{Z.~Liu},
\newblock \bibinfo{title}{A {{Data-driven Koopman Modeling Framework With
  Application}} to {{Soft Robots}}},
\newblock \bibinfo{journal}{International Journal of Control, Automation and
  Systems} \bibinfo{volume}{23} (\bibinfo{year}{2025})
  \bibinfo{pages}{249--261}.
\bibitem[{Bruder et~al.(2021)Bruder, Fu, Gillespie, Remy, and
  Vasudevan}]{9277915}
\bibinfo{author}{D.~Bruder}, \bibinfo{author}{X.~Fu}, \bibinfo{author}{R.~B.
  Gillespie}, \bibinfo{author}{C.~D. Remy}, \bibinfo{author}{R.~Vasudevan},
\newblock \bibinfo{title}{Data-driven control of soft robots using {Koopman}
  operator theory},
\newblock \bibinfo{journal}{IEEE Transactions on Robotics} \bibinfo{volume}{37}
  (\bibinfo{year}{2021}) \bibinfo{pages}{948--961}.
\bibitem[{Wang et~al.(2023)Wang, Xu, Lai, Wang, Hu, Li, and
  Song}]{wangImprovedKoopmanMPCFramework2023}
\bibinfo{author}{J.~Wang}, \bibinfo{author}{B.~Xu}, \bibinfo{author}{J.~Lai},
  \bibinfo{author}{Y.~Wang}, \bibinfo{author}{C.~Hu}, \bibinfo{author}{H.~Li},
  \bibinfo{author}{A.~Song},
\newblock \bibinfo{title}{An improved koopman-mpc framework for data-driven
  modeling and control of soft actuators},
\newblock \bibinfo{journal}{IEEE Robotics and Automation Letters}
  \bibinfo{volume}{8} (\bibinfo{year}{2023}) \bibinfo{pages}{616--623}.
\bibitem[{Lu et~al.(2024)Lu, Jiang, and Bai}]{luDeepEmbeddingKoopman2024}
\bibinfo{author}{J.~Lu}, \bibinfo{author}{J.~Jiang}, \bibinfo{author}{Y.~Bai},
\newblock \bibinfo{title}{Deep {{Embedding Koopman Neural Operator-Based
  Nonlinear Flight Training Trajectory Prediction Approach}}},
\newblock \bibinfo{journal}{Mathematics} \bibinfo{volume}{12}
  (\bibinfo{year}{2024}) \bibinfo{pages}{2162}.
\bibitem[{Lu et~al.(2025)Lu, Jiang, Bai, Dai, and
  Zhang}]{luFlightKoopmanDeepKoopman2025}
\bibinfo{author}{J.~Lu}, \bibinfo{author}{J.~Jiang}, \bibinfo{author}{Y.~Bai},
  \bibinfo{author}{W.~Dai}, \bibinfo{author}{W.~Zhang},
\newblock \bibinfo{title}{{{FlightKoopman}}: {{Deep Koopman}} for
  {{Multi-Dimensional Flight Trajectory Prediction}}},
\newblock \bibinfo{journal}{International Journal of Computational Intelligence
  and Applications}  (\bibinfo{year}{2025}) \bibinfo{pages}{2450038}.
\bibitem[{Li et~al.(2017)Li, Dietrich, Bollt, and
  Kevrekidis}]{liExtendedDynamicMode2017b}
\bibinfo{author}{Q.~Li}, \bibinfo{author}{F.~Dietrich}, \bibinfo{author}{E.~M.
  Bollt}, \bibinfo{author}{I.~G. Kevrekidis},
\newblock \bibinfo{title}{Extended dynamic mode decomposition with dictionary
  learning: A data-driven adaptive spectral decomposition of the {Koopman}
  operator},
\newblock \bibinfo{journal}{Chaos} \bibinfo{volume}{27} (\bibinfo{year}{2017})
  \bibinfo{pages}{103111}.
\bibitem[{Maksakov et~al.(2023)Maksakov, Golovin, Shysh, and
  Palis}]{MAKSAKOV2023110368}
\bibinfo{author}{A.~Maksakov}, \bibinfo{author}{I.~Golovin},
  \bibinfo{author}{M.~Shysh}, \bibinfo{author}{S.~Palis},
\newblock \bibinfo{title}{Data-driven modeling for damping and positioning
  control of gantry crane},
\newblock \bibinfo{journal}{Mechanical Systems and Signal Processing}
  \bibinfo{volume}{197} (\bibinfo{year}{2023}) \bibinfo{pages}{110368}.
\bibitem[{Leask et~al.(2021)Leask, McDonell, and
  Samuelsen}]{leaskModalExtractionSpatiotemporal2021a}
\bibinfo{author}{S.~B. Leask}, \bibinfo{author}{V.~G. McDonell},
  \bibinfo{author}{S.~Samuelsen},
\newblock \bibinfo{title}{Modal extraction of spatiotemporal atomization data
  using a deep convolutional {{Koopman}} network},
\newblock \bibinfo{journal}{Physics of Fluids} \bibinfo{volume}{33}
  (\bibinfo{year}{2021}) \bibinfo{pages}{033323}.
\bibitem[{{Alford-Lago} et~al.(2022){Alford-Lago}, Curtis, Ihler, and
  Issan}]{alford-lagoDeepLearningEnhanced2022}
\bibinfo{author}{D.~J. {Alford-Lago}}, \bibinfo{author}{C.~W. Curtis},
  \bibinfo{author}{A.~T. Ihler}, \bibinfo{author}{O.~Issan},
\newblock \bibinfo{title}{Deep learning enhanced dynamic mode decomposition},
\newblock \bibinfo{journal}{Chaos} \bibinfo{volume}{32} (\bibinfo{year}{2022})
  \bibinfo{pages}{033116}.
\bibitem[{Tang et~al.(2023)Tang, Zhu, and Zhang}]{9968056}
\bibinfo{author}{Y.~Tang}, \bibinfo{author}{Z.~Zhu},
  \bibinfo{author}{H.~Zhang},
\newblock \bibinfo{title}{A reachability-based spatio-temporal sampling
  strategy for kinodynamic motion planning},
\newblock \bibinfo{journal}{IEEE Robotics and Automation Letters}
  \bibinfo{volume}{8} (\bibinfo{year}{2023}) \bibinfo{pages}{448--455}.
\bibitem[{Wang et~al.(2024)Wang, Cao, Chen, and
  Kang}]{wangPhysicsinformedDeepKoopman2024}
\bibinfo{author}{X.~Wang}, \bibinfo{author}{Y.~Cao}, \bibinfo{author}{S.~Chen},
  \bibinfo{author}{Y.~Kang},
\newblock \bibinfo{title}{Physics-informed deep {{Koopman}} operator for
  {{Lagrangian}} dynamic systems},
\newblock \bibinfo{journal}{Science China Information Sciences}
  \bibinfo{volume}{67} (\bibinfo{year}{2024}) \bibinfo{pages}{192201}.
\bibitem[{Cranmer et~al.(2020)Cranmer, Greydanus, Hoyer, Battaglia, Spergel,
  and Ho}]{cranmer2020lagrangianneuralnetworks}
\bibinfo{author}{M.~Cranmer}, \bibinfo{author}{S.~Greydanus},
  \bibinfo{author}{S.~Hoyer}, \bibinfo{author}{P.~Battaglia},
  \bibinfo{author}{D.~Spergel}, \bibinfo{author}{S.~Ho},
  \bibinfo{title}{Lagrangian neural networks}, \bibinfo{year}{2020}.
  \DOIprefix\doi{https://arxiv.org/abs/2003.04630}.
  \href{http://arxiv.org/abs/arXiv: 2003.04630}{\tt arXiv:arXiv: 2003.04630}.
\bibitem[{Zhang et~al.(2024)Zhang, Zhu, and
  Lin}]{zhangLearningHamiltonianNeural2024}
\bibinfo{author}{J.~Zhang}, \bibinfo{author}{Q.~Zhu}, \bibinfo{author}{W.~Lin},
  \bibinfo{title}{Learning {{Hamiltonian}} neural {{Koopman}} operator and
  simultaneously sustaining and discovering conservation law},
  \bibinfo{year}{2024}. \DOIprefix\doi{10.48550/arXiv.2406.02154}.
  \href{http://arxiv.org/abs/2406.02154}{\tt arXiv:2406.02154}.
\bibitem[{Meng et~al.(2024)Meng, Liu, Shi, Ma, Ren, and Meng}]{10682797}
\bibinfo{author}{F.~Meng}, \bibinfo{author}{J.~Liu}, \bibinfo{author}{H.~Shi},
  \bibinfo{author}{H.~Ma}, \bibinfo{author}{H.~Ren}, \bibinfo{author}{M.~Q.-H.
  Meng},
\newblock \bibinfo{title}{Online time-informed kinodynamic motion planning of
  nonlinear systems},
\newblock \bibinfo{journal}{IEEE Robotics and Automation Letters}
  \bibinfo{volume}{9} (\bibinfo{year}{2024}) \bibinfo{pages}{9589--9596}.
\bibitem[{Tayal et~al.(2023)Tayal, Renganathan, Ghosh, Jia, and
  Kumar}]{tayal2023koopman}
\bibinfo{author}{K.~Tayal}, \bibinfo{author}{A.~Renganathan},
  \bibinfo{author}{R.~Ghosh}, \bibinfo{author}{X.~Jia},
  \bibinfo{author}{V.~Kumar},
\newblock \bibinfo{title}{Koopman invertible autoencoder: Leveraging forward
  and backward dynamics for temporal modeling},
\newblock in: \bibinfo{booktitle}{2023 IEEE International Conference on Data
  Mining (ICDM)}, \bibinfo{year}{2023}, pp. \bibinfo{pages}{588--597}.
  \DOIprefix\doi{10.1109/ICDM58522.2023.00068}.
\bibitem[{Azencot et~al.(2020)Azencot, Erichson, Lin, and
  Mahoney}]{azencot2020forecasting}
\bibinfo{author}{O.~Azencot}, \bibinfo{author}{N.~B. Erichson},
  \bibinfo{author}{V.~Lin}, \bibinfo{author}{M.~W. Mahoney},
  \bibinfo{title}{Forecasting sequential data using consistent koopman
  autoencoders}, \bibinfo{year}{2020}. \DOIprefix\doi{arXiv:2003.02236}.
\bibitem[{Jin et~al.(2023)Jin, Hou, Zhong, Yi, and Chen}]{JIN2023110604}
\bibinfo{author}{Y.~Jin}, \bibinfo{author}{L.~Hou}, \bibinfo{author}{S.~Zhong},
  \bibinfo{author}{H.~Yi}, \bibinfo{author}{Y.~Chen},
\newblock \bibinfo{title}{{Invertible Koopman Network} and its application in
  data-driven modeling for dynamic systems},
\newblock \bibinfo{journal}{Mechanical Systems and Signal Processing}
  \bibinfo{volume}{200} (\bibinfo{year}{2023}) \bibinfo{pages}{110604}.
\bibitem[{Jin et~al.(2024)Jin, Hou, and Zhong}]{jinExtendedDynamicMode2024}
\bibinfo{author}{Y.~Jin}, \bibinfo{author}{L.~Hou}, \bibinfo{author}{S.~Zhong},
\newblock \bibinfo{title}{Extended {{Dynamic Mode Decomposition}} with
  {{Invertible Dictionary Learning}}},
\newblock \bibinfo{journal}{Neural Networks} \bibinfo{volume}{173}
  (\bibinfo{year}{2024}) \bibinfo{pages}{106177}.
\bibitem[{Hou et~al.(2024)Hou, Zhang, and
  Fang}]{houInvertibleNeuralNetwork2024a}
\bibinfo{author}{X.~Hou}, \bibinfo{author}{J.~Zhang},
  \bibinfo{author}{L.~Fang},
\newblock \bibinfo{title}{Invertible neural network combined with dynamic mode
  decomposition applied to flow field feature extraction and prediction},
\newblock \bibinfo{journal}{Physics of Fluids} \bibinfo{volume}{36}
  (\bibinfo{year}{2024}) \bibinfo{pages}{095174}.
\bibitem[{Meng et~al.(2024)Meng, Huang, and Qiu}]{meng2023physicsinformed}
\bibinfo{author}{Y.~Meng}, \bibinfo{author}{J.~Huang},
  \bibinfo{author}{Y.~Qiu},
\newblock \bibinfo{title}{Koopman operator learning using invertible neural
  networks},
\newblock \bibinfo{journal}{Journal of Computational Physics}
  \bibinfo{volume}{501} (\bibinfo{year}{2024}) \bibinfo{pages}{112795}.
\bibitem[{Li et~al.(2023)Li, Liang, Lian, Liu, Zhu, and Liu}]{li2023invka}
\bibinfo{author}{F.~Li}, \bibinfo{author}{D.~Liang}, \bibinfo{author}{J.~Lian},
  \bibinfo{author}{Q.~Liu}, \bibinfo{author}{H.~Zhu}, \bibinfo{author}{J.~Liu},
  \bibinfo{title}{{InvKA}: Gait recognition via invertible koopman
  autoencoder}, \bibinfo{year}{2023}. \DOIprefix\doi{arXiv:2309.14764}.
\bibitem[{Curtis et~al.(2023)Curtis, {Jay Alford-Lago}, Bollt, and
  Tuma}]{curtisMachineLearningEnhanced2023}
\bibinfo{author}{C.~W. Curtis}, \bibinfo{author}{D.~{Jay Alford-Lago}},
  \bibinfo{author}{E.~Bollt}, \bibinfo{author}{A.~Tuma},
\newblock \bibinfo{title}{Machine learning enhanced {{Hankel}} dynamic-mode
  decomposition},
\newblock \bibinfo{journal}{Chaos: An Interdisciplinary Journal of Nonlinear
  Science} \bibinfo{volume}{33} (\bibinfo{year}{2023}) \bibinfo{pages}{083133}.
\bibitem[{Korda and Mezi{\'c}(2018)}]{KORDA2018149}
\bibinfo{author}{M.~Korda}, \bibinfo{author}{I.~Mezi{\'c}},
\newblock \bibinfo{title}{Linear predictors for nonlinear dynamical systems:
  {Koopman} operator meets model predictive control},
\newblock \bibinfo{journal}{Automatica} \bibinfo{volume}{93}
  (\bibinfo{year}{2018}) \bibinfo{pages}{149--160}.
\bibitem[{Lu et~al.(2021)Lu, Jin, Pang, Zhang, and Karniadakis}]{Lu_2021}
\bibinfo{author}{L.~Lu}, \bibinfo{author}{P.~Jin}, \bibinfo{author}{G.~Pang},
  \bibinfo{author}{Z.~Zhang}, \bibinfo{author}{G.~E. Karniadakis},
\newblock \bibinfo{title}{Learning nonlinear operators via {DeepONet} based on
  the universal approximation theorem of operators},
\newblock \bibinfo{journal}{Nature Machine Intelligence} \bibinfo{volume}{3}
  (\bibinfo{year}{2021}) \bibinfo{pages}{218--229}.
\bibitem[{Li et~al.(2025)Li, Miao, Khodaei, and Aliabadi}]{LI2025128675}
\bibinfo{author}{H.~Li}, \bibinfo{author}{Y.~Miao}, \bibinfo{author}{Z.~S.
  Khodaei}, \bibinfo{author}{M.~Aliabadi},
\newblock \bibinfo{title}{An architectural analysis of {DeepOnet} and a general
  extension of the physics-informed deeponet model on solving nonlinear
  parametric partial differential equations},
\newblock \bibinfo{journal}{Neurocomputing} \bibinfo{volume}{611}
  (\bibinfo{year}{2025}) \bibinfo{pages}{128675}.
\bibitem[{Chen et~al.(2024)Chen, Wang, Li, and
  Fu}]{chenHybridDecoderDeepONetOperator2024}
\bibinfo{author}{B.~Chen}, \bibinfo{author}{C.~Wang}, \bibinfo{author}{W.~Li},
  \bibinfo{author}{H.~Fu},
\newblock \bibinfo{title}{A hybrid {{Decoder-DeepONet}} operator regression
  framework for unaligned observation data},
\newblock \bibinfo{journal}{Physics of Fluids} \bibinfo{volume}{36}
  (\bibinfo{year}{2024}) \bibinfo{pages}{027132}.
\bibitem[{Venturi and Casey(2023)}]{venturiSVDPerspectivesAugmenting2023}
\bibinfo{author}{S.~Venturi}, \bibinfo{author}{T.~Casey},
\newblock \bibinfo{title}{{{SVD Perspectives}} for {{Augmenting DeepONet
  Flexibility}} and {{Interpretability}}},
\newblock \bibinfo{journal}{Computer Methods in Applied Mechanics and
  Engineering} \bibinfo{volume}{403} (\bibinfo{year}{2023})
  \bibinfo{pages}{115718}.
\bibitem[{Goswami et~al.(2024)Goswami, Jagtap, Babaee, Susi, and
  Karniadakis}]{GOSWAMI2024116674}
\bibinfo{author}{S.~Goswami}, \bibinfo{author}{A.~D. Jagtap},
  \bibinfo{author}{H.~Babaee}, \bibinfo{author}{B.~T. Susi},
  \bibinfo{author}{G.~E. Karniadakis},
\newblock \bibinfo{title}{Learning stiff chemical kinetics using extended deep
  neural operators},
\newblock \bibinfo{journal}{Computer Methods in Applied Mechanics and
  Engineering} \bibinfo{volume}{419} (\bibinfo{year}{2024})
  \bibinfo{pages}{116674}.
\bibitem[{Lee and Shin(2024)}]{doi:10.1137/23M1598751}
\bibinfo{author}{S.~Lee}, \bibinfo{author}{Y.~Shin},
\newblock \bibinfo{title}{On the training and generalization of deep operator
  networks},
\newblock \bibinfo{journal}{SIAM Journal on Scientific Computing}
  \bibinfo{volume}{46} (\bibinfo{year}{2024}) \bibinfo{pages}{C273--C296}.
\bibitem[{Li et~al.(2020{\natexlab{a}})Li, Kovachki, Azizzadenesheli, Liu,
  Bhattacharya, Stuart, and Anandkumar}]{li2020multipolegraphneuraloperator}
\bibinfo{author}{Z.~Li}, \bibinfo{author}{N.~Kovachki},
  \bibinfo{author}{K.~Azizzadenesheli}, \bibinfo{author}{B.~Liu},
  \bibinfo{author}{K.~Bhattacharya}, \bibinfo{author}{A.~Stuart},
  \bibinfo{author}{A.~Anandkumar}, \bibinfo{title}{Multipole graph neural
  operator for parametric partial differential equations},
  \bibinfo{year}{2020}{\natexlab{a}}.
  \DOIprefix\doi{https://arxiv.org/abs/2006.09535}.
  \href{http://arxiv.org/abs/arXiv:2006.09535}{\tt arXiv:arXiv:2006.09535}.
\bibitem[{Li et~al.(2020{\natexlab{b}})Li, Kovachki, Azizzadenesheli, Liu,
  Bhattacharya, Stuart, and Anandkumar}]{li2020neuraloperatorgraphkernel}
\bibinfo{author}{Z.~Li}, \bibinfo{author}{N.~Kovachki},
  \bibinfo{author}{K.~Azizzadenesheli}, \bibinfo{author}{B.~Liu},
  \bibinfo{author}{K.~Bhattacharya}, \bibinfo{author}{A.~Stuart},
  \bibinfo{author}{A.~Anandkumar}, \bibinfo{title}{Neural operator: Graph
  kernel network for partial differential equations},
  \bibinfo{year}{2020}{\natexlab{b}}.
  \DOIprefix\doi{https://arxiv.org/abs/2003.03485}.
  \href{http://arxiv.org/abs/arXiv:2003.03485}{\tt arXiv:arXiv:2003.03485}.
\bibitem[{Cao et~al.(2024)Cao, Goswami, and
  Karniadakis}]{caoLaplaceNeuralOperator2024}
\bibinfo{author}{Q.~Cao}, \bibinfo{author}{S.~Goswami}, \bibinfo{author}{G.~E.
  Karniadakis},
\newblock \bibinfo{title}{Laplace neural operator for solving differential
  equations},
\newblock \bibinfo{journal}{Nature Machine Intelligence} \bibinfo{volume}{6}
  (\bibinfo{year}{2024}) \bibinfo{pages}{631--640}.
\bibitem[{Li et~al.(2024)Li, Lai, Zhang, and
  Wang}]{li2024m2nomultiresolutionoperatorlearning}
\bibinfo{author}{Z.~Li}, \bibinfo{author}{Z.~Lai}, \bibinfo{author}{X.~Zhang},
  \bibinfo{author}{W.~Wang}, \bibinfo{title}{{M2NO}: Multiresolution operator
  learning with multiwavelet-based algebraic multigrid method},
  \bibinfo{year}{2024}. \DOIprefix\doi{https://arxiv.org/abs/2406.04822}.
  \href{http://arxiv.org/abs/arXiv:2406.04822}{\tt arXiv:arXiv:2406.04822}.
\bibitem[{Gupta et~al.(2021)Gupta, Xiao, and
  Bogdan}]{guptaMultiwaveletbasedOperatorLearning2021}
\bibinfo{author}{G.~Gupta}, \bibinfo{author}{X.~Xiao},
  \bibinfo{author}{P.~Bogdan}, \bibinfo{title}{Multiwavelet-based {{Operator
  Learning}} for {{Differential Equations}}}, \bibinfo{year}{2021}.
  \DOIprefix\doi{10.48550/arXiv.2109.13459}. \href{http://arxiv.org/abs/arXiv:
  2109.13459}{\tt arXiv:arXiv: 2109.13459}.
\bibitem[{Tripura and Chakraborty(2023)}]{TRIPURA2023115783}
\bibinfo{author}{T.~Tripura}, \bibinfo{author}{S.~Chakraborty},
\newblock \bibinfo{title}{Wavelet neural operator for solving parametric
  partial differential equations in computational mechanics problems},
\newblock \bibinfo{journal}{Computer Methods in Applied Mechanics and
  Engineering} \bibinfo{volume}{404} (\bibinfo{year}{2023})
  \bibinfo{pages}{115783}.
\bibitem[{Raoni{\'c} et~al.(2023)Raoni{\'c}, Molinaro, Ryck, Rohner,
  Bartolucci, Alaifari, Mishra, and
  de~B{\'e}zenac}]{raonicConvolutionalNeuralOperators2023b}
\bibinfo{author}{B.~Raoni{\'c}}, \bibinfo{author}{R.~Molinaro},
  \bibinfo{author}{T.~D. Ryck}, \bibinfo{author}{T.~Rohner},
  \bibinfo{author}{F.~Bartolucci}, \bibinfo{author}{R.~Alaifari},
  \bibinfo{author}{S.~Mishra}, \bibinfo{author}{E.~de~B{\'e}zenac},
  \bibinfo{title}{Convolutional {{Neural Operators}} for robust and accurate
  learning of {{PDEs}}}, \bibinfo{year}{2023}.
  \DOIprefix\doi{https://arxiv.org/abs/2302.01178}.
  \href{http://arxiv.org/abs/arXiv: 2302.01178}{\tt arXiv:arXiv: 2302.01178}.
\bibitem[{Cao(2021)}]{cao2021choosetransformerfouriergalerkin}
\bibinfo{author}{S.~Cao}, \bibinfo{title}{Choose a transformer: Fourier or
  galerkin}, \bibinfo{year}{2021}.
  \DOIprefix\doi{https://arxiv.org/abs/2105.14995}.
  \href{http://arxiv.org/abs/arXiv: 2105.14995}{\tt arXiv:arXiv: 2105.14995}.
\bibitem[{Li et~al.(2021)Li, Kovachki, Azizzadenesheli, Liu, Bhattacharya,
  Stuart, and Anandkumar}]{li2021fourierneuraloperatorparametric}
\bibinfo{author}{Z.~Li}, \bibinfo{author}{N.~Kovachki},
  \bibinfo{author}{K.~Azizzadenesheli}, \bibinfo{author}{B.~Liu},
  \bibinfo{author}{K.~Bhattacharya}, \bibinfo{author}{A.~Stuart},
  \bibinfo{author}{A.~Anandkumar}, \bibinfo{title}{Fourier neural operator for
  parametric partial differential equations}, \bibinfo{year}{2021}.
  \DOIprefix\doi{https://arxiv.org/abs/2010.08895}.
  \href{http://arxiv.org/abs/arXiv: 2010.08895}{\tt arXiv:arXiv: 2010.08895}.
\bibitem[{Guibas et~al.(2022)Guibas, Mardani, Li, Tao, Anandkumar, and
  Catanzaro}]{guibasAdaptiveFourierNeural2022}
\bibinfo{author}{J.~Guibas}, \bibinfo{author}{M.~Mardani},
  \bibinfo{author}{Z.~Li}, \bibinfo{author}{A.~Tao},
  \bibinfo{author}{A.~Anandkumar}, \bibinfo{author}{B.~Catanzaro},
  \bibinfo{title}{Adaptive {{Fourier Neural Operators}}: {{Efficient Token
  Mixers}} for {{Transformers}}}, \bibinfo{year}{2022}.
  \DOIprefix\doi{10.48550/arXiv.2111.13587}.
  \href{http://arxiv.org/abs/2111.13587}{\tt arXiv:2111.13587}.
\bibitem[{Pathak et~al.(2022)Pathak, Subramanian, Harrington, Raja,
  Chattopadhyay, Mardani, Kurth, Hall, Li, Azizzadenesheli, Hassanzadeh,
  Kashinath, and Anandkumar}]{pathakFourCastNetGlobalDatadriven2022}
\bibinfo{author}{J.~Pathak}, \bibinfo{author}{S.~Subramanian},
  \bibinfo{author}{P.~Harrington}, \bibinfo{author}{S.~Raja},
  \bibinfo{author}{A.~Chattopadhyay}, \bibinfo{author}{M.~Mardani},
  \bibinfo{author}{T.~Kurth}, \bibinfo{author}{D.~Hall},
  \bibinfo{author}{Z.~Li}, \bibinfo{author}{K.~Azizzadenesheli},
  \bibinfo{author}{P.~Hassanzadeh}, \bibinfo{author}{K.~Kashinath},
  \bibinfo{author}{A.~Anandkumar}, \bibinfo{title}{{{FourCastNet}}: {{A Global
  Data-driven High-resolution Weather Model}} using {{Adaptive Fourier Neural
  Operators}}}, \bibinfo{year}{2022}.
  \DOIprefix\doi{10.48550/arXiv.2202.11214}. \href{http://arxiv.org/abs/arXiv:
  2202.11214}{\tt arXiv:arXiv: 2202.11214}.
\bibitem[{Rahman et~al.(2023)Rahman, Ross, and
  Azizzadenesheli}]{rahmanUshapedNeuralOperators2023}
\bibinfo{author}{M.~A. Rahman}, \bibinfo{author}{Z.~E. Ross},
  \bibinfo{author}{K.~Azizzadenesheli}, \bibinfo{title}{U-{{NO}}: {{U-shaped
  Neural Operators}}}, \bibinfo{year}{2023}.
  \DOIprefix\doi{10.48550/arXiv.2204.11127}. \href{http://arxiv.org/abs/arXiv:
  2204.11127}{\tt arXiv:arXiv: 2204.11127}.
\bibitem[{Wen et~al.(2022)Wen, Li, Azizzadenesheli, Anandkumar, and
  Benson}]{WEN2022104180}
\bibinfo{author}{G.~Wen}, \bibinfo{author}{Z.~Li},
  \bibinfo{author}{K.~Azizzadenesheli}, \bibinfo{author}{A.~Anandkumar},
  \bibinfo{author}{S.~M. Benson},
\newblock \bibinfo{title}{{U-FNO}-an enhanced fourier neural operator-based
  deep-learning model for multiphase flow},
\newblock \bibinfo{journal}{Advances in Water Resources} \bibinfo{volume}{163}
  (\bibinfo{year}{2022}) \bibinfo{pages}{104180}.
\bibitem[{You et~al.(2022)You, Zhang, Ross, Lee, and
  Yu}]{youLearningDeepImplicit2022}
\bibinfo{author}{H.~You}, \bibinfo{author}{Q.~Zhang}, \bibinfo{author}{C.~J.
  Ross}, \bibinfo{author}{C.-H. Lee}, \bibinfo{author}{Y.~Yu},
\newblock \bibinfo{title}{Learning deep {{Implicit Fourier Neural Operators}}
  ({{IFNOs}}) with applications to heterogeneous material modeling},
\newblock \bibinfo{journal}{Computer Methods in Applied Mechanics and
  Engineering} \bibinfo{volume}{398} (\bibinfo{year}{2022})
  \bibinfo{pages}{115296}.
\bibitem[{Meng et~al.(2023)Meng, Zhu, Wang, and
  Shi}]{mengFastFlowPrediction2023}
\bibinfo{author}{D.~Meng}, \bibinfo{author}{Y.~Zhu}, \bibinfo{author}{J.~Wang},
  \bibinfo{author}{Y.~Shi},
\newblock \bibinfo{title}{Fast flow prediction of airfoil dynamic stall based
  on {{Fourier}} neural operator},
\newblock \bibinfo{journal}{Physics of Fluids} \bibinfo{volume}{35}
  (\bibinfo{year}{2023}) \bibinfo{pages}{115126}.
\bibitem[{Peng et~al.(2024)Peng, Qin, Yang, Wang, Liu, and
  Wang}]{PENG2024111063}
\bibinfo{author}{W.~Peng}, \bibinfo{author}{S.~Qin}, \bibinfo{author}{S.~Yang},
  \bibinfo{author}{J.~Wang}, \bibinfo{author}{X.~Liu}, \bibinfo{author}{L.~L.
  Wang},
\newblock \bibinfo{title}{Fourier neural operator for real-time simulation of
  3d dynamic urban microclimate},
\newblock \bibinfo{journal}{Building and Environment} \bibinfo{volume}{248}
  (\bibinfo{year}{2024}) \bibinfo{pages}{111063}.
\bibitem[{Xiong et~al.(2024)Xiong, Huang, Zhang, Deng, Sun, and
  Tian}]{XIONG2024113194}
\bibinfo{author}{W.~Xiong}, \bibinfo{author}{X.~Huang},
  \bibinfo{author}{Z.~Zhang}, \bibinfo{author}{R.~Deng},
  \bibinfo{author}{P.~Sun}, \bibinfo{author}{Y.~Tian},
\newblock \bibinfo{title}{Koopman neural operator as a mesh-free solver of
  non-linear partial differential equations},
\newblock \bibinfo{journal}{Journal of Computational Physics}
  \bibinfo{volume}{513} (\bibinfo{year}{2024}) \bibinfo{pages}{113194}.
\bibitem[{Xiong et~al.(2023)Xiong, Ma, Huang, Zhang, Sun, and
  Tian}]{xiongKoopmanLabMachineLearning2023}
\bibinfo{author}{W.~Xiong}, \bibinfo{author}{M.~Ma},
  \bibinfo{author}{X.~Huang}, \bibinfo{author}{Z.~Zhang},
  \bibinfo{author}{P.~Sun}, \bibinfo{author}{Y.~Tian},
\newblock \bibinfo{title}{{{KoopmanLab}}: {{Machine}} learning for solving
  complex physics equations},
\newblock \bibinfo{journal}{{APL} Machine Learning} \bibinfo{volume}{1}
  (\bibinfo{year}{2023}) \bibinfo{pages}{036110}.
\bibitem[{Meng et~al.(2024)Meng, Zhu, Wang, and
  Shi}]{mengKoopmanNeuralOperator2024a}
\bibinfo{author}{D.~Meng}, \bibinfo{author}{Y.~Zhu}, \bibinfo{author}{J.~Wang},
  \bibinfo{author}{Y.~Shi},
\newblock \bibinfo{title}{Koopman neural operator approach to fast flow
  prediction of airfoil transonic buffet},
\newblock \bibinfo{journal}{Physics of Fluids} \bibinfo{volume}{36}
  (\bibinfo{year}{2024}) \bibinfo{pages}{075182}.
\bibitem[{Cao et~al.(2024)Cao, Xu, Huang, Lv, Zhang, and Ding}]{CAO2024544}
\bibinfo{author}{Z.~Cao}, \bibinfo{author}{W.~Xu}, \bibinfo{author}{T.~Huang},
  \bibinfo{author}{Y.~Lv}, \bibinfo{author}{X.-M. Zhang},
  \bibinfo{author}{H.~Ding},
\newblock \bibinfo{title}{An efficient surrogate model for prediction of stress
  released distortion in large blade machining},
\newblock \bibinfo{journal}{Journal of Manufacturing Processes}
  \bibinfo{volume}{132} (\bibinfo{year}{2024}) \bibinfo{pages}{544--557}.
\bibitem[{Williams et~al.(2015)Williams, Kevrekidis, and
  Rowley}]{williamsDataDrivenApproximation2015}
\bibinfo{author}{M.~O. Williams}, \bibinfo{author}{I.~G. Kevrekidis},
  \bibinfo{author}{C.~W. Rowley},
\newblock \bibinfo{title}{A data-driven approximation of the {Koopman}
  operator: {Extending Dynamic Mode Decomposition}},
\newblock \bibinfo{journal}{Journal of Nonlinear Science} \bibinfo{volume}{25}
  (\bibinfo{year}{2015}) \bibinfo{pages}{1307--1346}.
\bibitem[{Garmaev and Fink(2024)}]{GARMAEV2024110351}
\bibinfo{author}{S.~Garmaev}, \bibinfo{author}{O.~Fink},
\newblock \bibinfo{title}{Deep koopman operator-based degradation modelling},
\newblock \bibinfo{journal}{Reliability Engineering \& System Safety}
  \bibinfo{volume}{251} (\bibinfo{year}{2024}) \bibinfo{pages}{110351}.
\bibitem[{Yu et~al.(2025)Yu, Zhou, Huang, Zhang, and
  Wang}]{yuDynamicModeDecomposition2025a}
\bibinfo{author}{Y.~Yu}, \bibinfo{author}{H.~Zhou}, \bibinfo{author}{B.~Huang},
  \bibinfo{author}{F.~Zhang}, \bibinfo{author}{B.~Wang},
\newblock \bibinfo{title}{Dynamic mode decomposition and short-time prediction
  of {{PM}}{\textsubscript{2.5}} using the graph {{Neural Koopman}} network},
\newblock \bibinfo{journal}{International Journal of Geographical Information
  Science} \bibinfo{volume}{39} (\bibinfo{year}{2025})
  \bibinfo{pages}{277--300}.
\bibitem[{Takens(1981)}]{10.1007/BFb0091924}
\bibinfo{author}{F.~Takens},
\newblock \bibinfo{title}{Detecting strange attractors in turbulence},
\newblock in: \bibinfo{editor}{D.~Rand}, \bibinfo{editor}{L.-S. Young} (Eds.),
  \bibinfo{booktitle}{Dynamical Systems and Turbulence, Warwick 1980},
  \bibinfo{publisher}{Springer Berlin Heidelberg}, \bibinfo{address}{Berlin,
  Heidelberg}, \bibinfo{year}{1981}, pp. \bibinfo{pages}{366--381}.
\bibitem[{Brunton et~al.(2017)Brunton, Brunton, Proctor, Kaiser, and
  Kutz}]{bruntonChaosIntermittentlyForced2017}
\bibinfo{author}{S.~L. Brunton}, \bibinfo{author}{B.~W. Brunton},
  \bibinfo{author}{J.~L. Proctor}, \bibinfo{author}{E.~Kaiser},
  \bibinfo{author}{J.~N. Kutz},
\newblock \bibinfo{title}{Chaos as an intermittently forced linear system},
\newblock \bibinfo{journal}{Nature Communications} \bibinfo{volume}{8}
  (\bibinfo{year}{2017}) \bibinfo{pages}{19}.
\bibitem[{Jacobsen et~al.(2018)Jacobsen, Smeulders, and
  Oyallon}]{jacobsen2018irevnet}
\bibinfo{author}{J.-H. Jacobsen}, \bibinfo{author}{A.~Smeulders},
  \bibinfo{author}{E.~Oyallon}, \bibinfo{title}{{i-RevNet}: Deep invertible
  networks}, \bibinfo{year}{2018}. \DOIprefix\doi{arXiv:1802.07088}.
\bibitem[{Papamakarios et~al.(2021)Papamakarios, Nalisnick, Rezende, Mohamed,
  and Lakshminarayanan}]{papamakarios2021normalizing}
\bibinfo{author}{G.~Papamakarios}, \bibinfo{author}{E.~Nalisnick},
  \bibinfo{author}{D.~J. Rezende}, \bibinfo{author}{S.~Mohamed},
  \bibinfo{author}{B.~Lakshminarayanan}, \bibinfo{title}{Normalizing flows for
  probabilistic modeling and inference}, \bibinfo{year}{2021}.
  \DOIprefix\doi{https://arxiv.org/abs/1912.02762}.
  \href{http://arxiv.org/abs/arXiv: 1912.02762}{\tt arXiv:arXiv: 1912.02762}.
\bibitem[{Gomez et~al.(2017)Gomez, Ren, Urtasun, and
  Grosse}]{gomez2017reversible}
\bibinfo{author}{A.~N. Gomez}, \bibinfo{author}{M.~Ren},
  \bibinfo{author}{R.~Urtasun}, \bibinfo{author}{R.~B. Grosse},
  \bibinfo{title}{The reversible residual network: Backpropagation without
  storing activations}, \bibinfo{year}{2017}.
  \DOIprefix\doi{https://arxiv.org/abs/1707.04585}.
  \href{http://arxiv.org/abs/arXiv: 1707.04585}{\tt arXiv:arXiv: 1707.04585}.
\bibitem[{Park and Kim(2022)}]{park2022visiontransformerswork}
\bibinfo{author}{N.~Park}, \bibinfo{author}{S.~Kim}, \bibinfo{title}{How do
  vision transformers work?}, \bibinfo{year}{2022}.
  \DOIprefix\doi{https://arxiv.org/abs/2202.06709}.
  \href{http://arxiv.org/abs/arXiv: 2202.06709}{\tt arXiv:arXiv: 2202.06709}.
\bibitem[{Kingma and Ba(2017)}]{kingma2017adam}
\bibinfo{author}{D.~P. Kingma}, \bibinfo{author}{J.~Ba}, \bibinfo{title}{Adam:
  A method for stochastic optimization}, \bibinfo{year}{2017}.
  \DOIprefix\doi{arXiv:1412.6980}.
\bibitem[{Takamoto et~al.(2024)Takamoto, Praditia, Leiteritz, MacKinlay,
  Alesiani, Pflüger, and
  Niepert}]{takamoto2024pdebenchextensivebenchmarkscientific}
\bibinfo{author}{M.~Takamoto}, \bibinfo{author}{T.~Praditia},
  \bibinfo{author}{R.~Leiteritz}, \bibinfo{author}{D.~MacKinlay},
  \bibinfo{author}{F.~Alesiani}, \bibinfo{author}{D.~Pflüger},
  \bibinfo{author}{M.~Niepert}, \bibinfo{title}{{PDEBENCH}: An extensive
  benchmark for scientific machine learning}, \bibinfo{year}{2024}.
  \DOIprefix\doi{https://arxiv.org/abs/2210.07182}.
  \href{http://arxiv.org/abs/arXiv: 2210.07182}{\tt arXiv:arXiv: 2210.07182}.
\bibitem[{Lu et~al.(2022)Lu, Meng, Cai, Mao, Goswami, Zhang, and
  Karniadakis}]{luComprehensiveFairComparison2022}
\bibinfo{author}{L.~Lu}, \bibinfo{author}{X.~Meng}, \bibinfo{author}{S.~Cai},
  \bibinfo{author}{Z.~Mao}, \bibinfo{author}{S.~Goswami},
  \bibinfo{author}{Z.~Zhang}, \bibinfo{author}{G.~E. Karniadakis},
\newblock \bibinfo{title}{A comprehensive and fair comparison of two neural
  operators (with practical extensions) based on {{FAIR}} data},
\newblock \bibinfo{journal}{Computer Methods in Applied Mechanics and
  Engineering} \bibinfo{volume}{393} (\bibinfo{year}{2022})
  \bibinfo{pages}{114778}.
\bibitem[{Lye et~al.(2021)Lye, Mishra, Ray, and Chandrashekar}]{LYE2021113575}
\bibinfo{author}{K.~O. Lye}, \bibinfo{author}{S.~Mishra},
  \bibinfo{author}{D.~Ray}, \bibinfo{author}{P.~Chandrashekar},
\newblock \bibinfo{title}{Iterative surrogate model optimization (ismo): An
  active learning algorithm for pde constrained optimization with deep neural
  networks},
\newblock \bibinfo{journal}{Computer Methods in Applied Mechanics and
  Engineering} \bibinfo{volume}{374} (\bibinfo{year}{2021})
  \bibinfo{pages}{113575}.
\bibitem[{Li et~al.(2023)Li, Huang, Liu, and Anandkumar}]{JMLR:v24:23-0064}
\bibinfo{author}{Z.~Li}, \bibinfo{author}{D.~Z. Huang},
  \bibinfo{author}{B.~Liu}, \bibinfo{author}{A.~Anandkumar},
\newblock \bibinfo{title}{Fourier neural operator with learned deformations for
  pdes on general geometries},
\newblock \bibinfo{journal}{Journal of Machine Learning Research}
  \bibinfo{volume}{24} (\bibinfo{year}{2023}) \bibinfo{pages}{1--26}.
\bibitem[{ECMWF(2 25)}]{ERA5_url}
\bibinfo{author}{ECMWF}, \bibinfo{title}{{ECMWF Reanalysis v5 (ERA5)}},
  \bibinfo{year}{Accessed: 2024-12-25}.
  \DOIprefix\doi{https://www.ecmwf.int/en/forecasts/dataset/ecmwf-reanalysis-v5}.
\bibitem[{Tran et~al.(2023)Tran, Mathews, Xie, and
  Ong}]{tranFactorizedFourierNeural2023}
\bibinfo{author}{A.~Tran}, \bibinfo{author}{A.~Mathews},
  \bibinfo{author}{L.~Xie}, \bibinfo{author}{C.~S. Ong},
  \bibinfo{title}{Factorized {{Fourier Neural Operators}}},
  \bibinfo{year}{2023}. \DOIprefix\doi{10.48550/arXiv.2111.13802}.
  \href{http://arxiv.org/abs/arXiv: 2111.13802}{\tt arXiv:arXiv: 2111.13802}.
\bibitem[{Liu et~al.(2025)Liu, Liu, Zhang, and
  Liu}]{liuMSIUFFNOMultiscaleImplicit2025}
\bibinfo{author}{S.~Liu}, \bibinfo{author}{H.~Liu}, \bibinfo{author}{T.~Zhang},
  \bibinfo{author}{X.~Liu},
\newblock \bibinfo{title}{{{MS-IUFFNO}}: {{Multi-scale}} implicit {{U-net}}
  enhanced factorized fourier neural operator for solving geometric {{PDEs}}},
\newblock \bibinfo{journal}{Computer Methods in Applied Mechanics and
  Engineering} \bibinfo{volume}{437} (\bibinfo{year}{2025})
  \bibinfo{pages}{117761}.
\bibitem[{Lingsch et~al.(2024)Lingsch, Michelis, de~Bezenac, Perera,
  Katzschmann, and Mishra}]{lingschRegularGridsFourierBased2024}
\bibinfo{author}{L.~Lingsch}, \bibinfo{author}{M.~Y. Michelis},
  \bibinfo{author}{E.~de~Bezenac}, \bibinfo{author}{S.~M. Perera},
  \bibinfo{author}{R.~K. Katzschmann}, \bibinfo{author}{S.~Mishra},
  \bibinfo{title}{Beyond {{Regular Grids}}: {{Fourier-Based Neural Operators}}
  on {{Arbitrary Domains}}}, \bibinfo{year}{2024}.
  \DOIprefix\doi{10.48550/arXiv.2305.19663}. \href{http://arxiv.org/abs/arXiv:
  2305.19663}{\tt arXiv:arXiv: 2305.19663}.
\bibitem[{Huang and Qiu(2025)}]{huangResolutionInvariantDeep2025a}
\bibinfo{author}{J.~Huang}, \bibinfo{author}{Y.~Qiu},
\newblock \bibinfo{title}{Resolution invariant deep operator network for
  {{PDEs}} with complex geometries},
\newblock \bibinfo{journal}{Journal of Computational Physics}
  \bibinfo{volume}{522} (\bibinfo{year}{2025}) \bibinfo{pages}{113601}.

\end{thebibliography}



\end{document}